\documentclass[10pt,twocolumn,letterpaper]{article}

\usepackage[pagenumbers]{cvpr} % To force page numbers, e.g. for an arXiv version
\usepackage{float}
\usepackage{multirow}
\usepackage{booktabs}
\usepackage[normalem]{ulem}
\usepackage{lineno} %
\definecolor{cvprblue}{rgb}{0.21,0.49,0.74}
\usepackage[pagebackref,breaklinks,colorlinks,allcolors=cvprblue]{hyperref}

\def\paperID{*****} % *** Enter the Paper ID here
\def\confName{CVPR}
\def\confYear{2026}

\title{E-3DPSM: A State Machine for Event-Based Egocentric \\ 3D Human Pose Estimation}

\author{Mayur Deshmukh$^1$$^,$$^2$\quad
Hiroyasu Akada$^1$\quad
Helge Rhodin$^1$$^,$$^3$\quad
Christian Theobalt$^1$\quad
Vladislav Golyanik$^1$\quad \\ \\
$^1$MPI for Informatics, SIC \quad
$^2$Saarland University, SIC \quad 
$^3$Bielefeld University
}

\begin{document}
% \maketitle
\newcommand{\mainresultstable}{
\begin{table}[t]
\caption{\textbf{Quantitative comparisons on EE3D-R and EE3D-W}. 
}
\vspace{-5pt}
\centering
\resizebox{\columnwidth}{!}{
\begin{tabular}{lccc}
\toprule
\textbf{EE3D-R} & \textbf{MPJPE} $\downarrow$ & \textbf{PA-MPJPE} $\downarrow$ & \textbf{$e_{\textbf{smooth}}$} $\downarrow$ \\
\midrule
EgoPoseFormer~\cite{yang2024egoposeformer} & 151.66 & 96.99 & 66.50 \\
EventEgo3D~\cite{Millerdurai2024CVPR} & 110.39 & 84.52 & 27.06 \\
EventEgo3D++~\cite{Millerdurai2025IJCV} & 103.28 & 77.06 & 22.93 \\
\midrule
\textbf{\textit{Ours (Causal)}} & \textbf{84.45} & \textbf{62.64} & \textbf{8.40} \\
\textbf{\textit{Ours (Non-Causal)}} & \textbf{81.32} & \textbf{60.21} & \textbf{6.65} \\
\midrule
\textbf{EE3D-W} & \textbf{MPJPE} $\downarrow$ & \textbf{PA-MPJPE} $\downarrow$ & \textbf{$e_{\textbf{smooth}}$} $\downarrow$ \\
\midrule
EgoPoseFormer~\cite{yang2024egoposeformer} & 220.40 & 130.45 & 79.23 \\
EventEgo3D~\cite{Millerdurai2024CVPR}  & 195.50 & 108.20 & 45.29 \\
EventEgo3D++~\cite{Millerdurai2025IJCV} & 172.43 & 98.41 & 40.87 \\
\midrule
\textbf{\textit{Ours (Causal)}} & \textbf{158.86} & \textbf{93.46} & \textbf{23.57} \\
\textbf{\textit{Ours (Non-Causal)}} & \textbf{155.82} & \textbf{90.85} & \textbf{22.65} \\
\bottomrule
\end{tabular}
}
\vspace{-10pt}
\label{tab:main_results}
\end{table}
}

\newcommand{\avgocclusiononlytable}{
\begin{table}[t]
\caption{\textbf{Occlusion-only quantitative comparison on EE3D-R and EE3D-W}. Evaluation is performed only on occluded joints.
% excluding visible ones. 
}
\centering
\renewcommand{\arraystretch}{0.8}
\setlength{\tabcolsep}{16pt}       % reduce column padding
\vspace{-3pt}
\resizebox{\columnwidth}{!}{
\begin{tabular}{lcc}
\toprule
\textbf{EE3D-R} & \textbf{MPJPE} $\downarrow$ & \textbf{PA-MPJPE} $\downarrow$ \\
\midrule
EgoPoseFormer~\cite{yang2024egoposeformer} & 110.18 & 55.64 \\
EventEgo3D~\cite{Millerdurai2024CVPR} & 86.90 & 50.93  \\
EventEgo3D++~\cite{Millerdurai2025IJCV} & 88.43 & 49.53  \\
\midrule
\textbf{\textit{Ours (Causal)}} & \textbf{67.49} & \textbf{41.85}  \\
\textbf{\textit{Ours (Non-Causal)}} & \textbf{64.69} & \textbf{40.10} \\
\midrule
\textbf{EE3D-W} & \textbf{MPJPE} $\downarrow$ & \textbf{PA-MPJPE} $\downarrow$ \\
\midrule
EgoPoseFormer~\cite{yang2024egoposeformer} & 173.01 & 113.67  \\
EventEgo3D~\cite{Millerdurai2024CVPR}  & 159.04 & 84.84 \\
EventEgo3D++~\cite{Millerdurai2025IJCV} & 147.42 & 77.64 \\
\midrule
\textbf{\textit{Ours (Causal)}} & \textbf{135.84} & \textbf{73.19} \\
\textbf{\textit{Ours (Non-Causal)}} & \textbf{134.72} & \textbf{72.29} \\
\bottomrule
\end{tabular}
}
\vspace{-10pt}
\label{tab:avg_occlusion_only_results}
\end{table}
}

\newcommand{\mainablationtable}{
\begin{table}[t]
\caption{\textbf{Ablation study on the EE3D-R dataset} evaluating the impact of each component of our \textit{E-3DPSM} approach. 
}
\centering
\resizebox{\columnwidth}{!}{
\begin{tabular}{lccc}
\toprule
\textbf{SPEM (Sec. \ref{sec:spem})} & \textbf{MPJPE} $\downarrow$ & \textbf{PA-MPJPE} $\downarrow$ & \textbf{$e_{\textbf{smooth}}$} $\downarrow$ \\
\midrule
w/o SSM Blocks                   & 118.53 & 87.48 & 16.94 \\
Single SSM Block (Stage 4)       &  90.18 & 66.06  & 7.84 \\
w/o Deformable Attention        &  88.27 & 64.98  & 7.71 \\
\midrule
\textbf{PRM (Sec. \ref{sec:prm})} & \textbf{MPJPE} $\downarrow$ & \textbf{PA-MPJPE} $\downarrow$ & \textbf{$e_{\textbf{smooth}}$} $\downarrow$ \\
\midrule
w/o Fusion Module                & 141.22 & 84.17 & 10.14 \\
Direct Pose Only               & 91.26 & 65.43 & 17.22 \\
Static Fusion (Non-learned)      & 88.31 & 65.63 & 9.93 \\
\midrule
\textbf{\textit{Ours (full model)}}  & \textbf{84.45} & \textbf{62.64} & \textbf{8.40} \\
\bottomrule
\end{tabular}
}
\vspace{-10pt}
\label{tab:main_ablations}
\end{table}
}

% -------------Supplementary Tables----------------------------------

\newcommand{\modelefficiencytable}{
\begin{table}[t]
\caption{\textbf{Model efficiency comparison} in terms of parameters, FLOPs, GPU memory, and 3D pose update rate in Hz (measured on a single NVIDIA A6000 GPU).}
\centering
\resizebox{\columnwidth}{!}{
\begin{tabular}{lcccc}
\toprule
\textbf{Method} & \textbf{Params} $\downarrow$ & \textbf{FLOPs} $\downarrow$ & \textbf{GPU Memory} $\downarrow$ & \textbf{Pose Update Rate} $\uparrow$ \\
\midrule
EgoPoseFormer \cite{yang2024egoposeformer} & 14.1 M & 5.5 G & 82 MB & 130 \\ 
EventEgo3D \cite{Millerdurai2024CVPR} & \textbf{1.25 M} & \textbf{416.84 M} & \textbf{25 MB} & \textbf{139} \\
EventEgo3D++ \cite{Millerdurai2025IJCV} & \textbf{1.25 M} & \textbf{416.84 M} & \textbf{25 MB} &\textbf{139} \\
\midrule
{\textbf{\textit{Ours}}} & 6.64 M & 8.16 G & 74 MB &80 \\
\bottomrule
\end{tabular}
}
\label{tab:model_efficiency}
\end{table}
}

\newcommand{\gpumemorydetailed}{
\begin{table}[t]
\caption{Detailed module-wise FLOPs breakdown.}
\centering
\begin{tabular}{lc}
\toprule
\textbf{Modules} & \textbf{FLOPs} \\
\midrule
CNN Layers           & 7.05 G \\
Deformable Attention & 0.89 G \\
SSM                  & 0.06 G \\
Query Decoder        & 0.15 G \\
Pose Heads           & $10^{-3}$ G \\
Fusion               & $10^{-4}$ G \\
\midrule
\textbf{\textit{Total}} & \textbf{8.16 G} \\
\bottomrule
\end{tabular}
\label{tab:gpumemorydetailed}
\end{table}
}

\newcommand{\qrablation}{
\begin{table}[t]
\caption{\textbf{Ablation of global and input/state-dependent covariance learning for Q and R on EE3D-R dataset.}}
\centering
\begin{tabular}{lcc}
\toprule
\textbf{Learning Strategy} & \textbf{MPJPE} $\downarrow$ & \textbf{PA-MPJPE} $\downarrow$ \\
\midrule
Input/State Dependent & 91.15 & 69.74 \\
\textit{Global Learned (Ours)} & 84.45 & 62.64 \\
\bottomrule
\end{tabular}
\label{tab:q_r_ablation}
\end{table}
}

\newcommand{\trainingablationtable}{
\begin{table}[t]
\caption{\textbf{Training strategy ablation on the EE3D-R dataset.} We compare causal (forward) vs.~non-causal (bidirectional) training and different sequence lengths used during training.}
\centering
\resizebox{\columnwidth}{!}{
\begin{tabular}{lccc}
\toprule
\textbf{Training Strategy} & \textbf{MPJPE $\downarrow$} & \textbf{PA-MPJPE $\downarrow$} & \textbf{$e_{\textbf{smooth}} \downarrow$}  \\
\midrule
\multicolumn{4}{l}{\textbf{Training Directionality}} \\
Causal (Forward Only)         & 89.88 &  66.74  & 10.14 \\
\midrule
\textbf{\textit{Non-Causal (Ours)}} & \textbf{84.45} & \textbf{62.64} & \textbf{8.40} \\
\midrule
\multicolumn{4}{l}{\textbf{Pose Sequence Length ($\boldsymbol{\mathbf{N}}$)}} \\
20 poses  & 86.25 & 65.62 & 9.76 \\
30 poses  &  86.03 & 64.87 & 8.95 \\
\midrule
\textbf{\textit{40 Poses (Ours)}}    & \textbf{84.45} & \textbf{62.64} & \textbf{8.40} \\
\bottomrule
\end{tabular}
}
\label{tab:training_ablation}
\end{table}
}

\newcommand{\stateresetablationtable}{
\begin{table}[t]
\caption{\textbf{Inference-time ablation on the EE3D-R dataset comparing different strategies for resetting internal states.} We evaluate resetting the SSM block states, resetting the Kalman fusion states, and using continuous state evolution without resets (ours).}
\centering
\resizebox{\columnwidth}{!}{
\begin{tabular}{lccc}
\toprule
\textbf{State Reset Strategy} & \textbf{MPJPE $\downarrow$} & \textbf{PA-MPJPE $\downarrow$} & \textbf{$e_{\textbf{smooth}} \downarrow$}  \\
\midrule
SSM Reset (40 Frames)      & 104.56 & 81.24   & 15.20 \\
Fusion Reset (40 Frames) &  95.40 & 70.90 & 21.50 \\
\midrule
\textbf{\textit{No Reset (Ours)}} &  \textbf{84.45} & \textbf{62.64} & \textbf{8.40} \\
\bottomrule
\end{tabular}
}
\label{tab:state_reset_ablation}
\end{table}
}

\newcommand{\resultswkftable}{
\begin{table}[t]
\caption{\textbf{Comparison with Kalman-smoothed baselines on the EE3D-R dataset.} 
We apply inference-time Kalman filtering (KF) to prior methods to rule out post-hoc smoothing as the main reason for improvements. 
Our method achieves substantially lower MPJPE and $e_{\text{smooth}}$, demonstrating that the performance is due to the proposed architecture and not filtering in post-processing.}
\centering
\resizebox{\columnwidth}{!}{
\begin{tabular}{lccc}
\toprule
\textbf{Method} & \textbf{MPJPE} $\downarrow$ & \textbf{PA-MPJPE} $\downarrow$ & \textbf{$e_{\textbf{smooth}}$} $\downarrow$ \\
\midrule
EgoPoseFormer~\cite{yang2024egoposeformer} with KF & 144.27 & 92.32 & 37.11 \\
EventEgo3D~\cite{Millerdurai2024CVPR} with KF    & 107.23 & 82.54 & 15.78 \\
EventEgo3D++~\cite{Millerdurai2025IJCV} with KF  & 100.98 & 75.57 & 13.98 \\
\midrule
\textbf{\textit{Ours (Causal)}}      & \textbf{84.45} & \textbf{62.64} & \textbf{8.40} \\ 
\textbf{\textit{Ours (Non-Causal)}}  & \textbf{81.32} & \textbf{60.21} & \textbf{6.65} \\
\bottomrule
\end{tabular}
}
\label{tab:main_results_w_kf}
\end{table}
}

\newcommand{\eventreprablationtable}{
\begin{table}[t]
\caption{\textbf{Design choice study for event stream representation for learning on the EE3D-R dataset.} We experiment with learned voxel-based representation, learned LNES and fixed LNES. 
}
\centering
\resizebox{\columnwidth}{!}{
\begin{tabular}{lccc}
\toprule
\textbf{Event Representation} & \textbf{MPJPE $\downarrow$} & \textbf{PA-MPJPE $\downarrow$} & \textbf{$e_{\textbf{smooth}} \downarrow$}  \\
\midrule
Learned Voxel-Based \cite{Gehrig2019} & 100.25 &  78.54   & 10.78 \\
Learned LNES &   93.78   & 71.12 & 9.14 \\
\midrule
\textbf{\textit{LNES (Ours)}} & \textbf{84.45} & \textbf{62.64} & \textbf{8.40} \\
\bottomrule
\end{tabular}
}
\label{tab:event_repr_ablation}
\end{table}
}

\newcommand{\eventfrequenciesablationtable}{
\begin{table}[t]
\caption{\textbf{Inference-time ablation on the EE3D-R dataset comparing the use of different
3D pose update rates 
% event 
frequencies.}}
\centering
\resizebox{\columnwidth}{!}{
\begin{tabular}{lccc}
\toprule
\textbf{Event Frequencies} & \textbf{MPJPE $\downarrow$} & \textbf{PA-MPJPE $\downarrow$} & \textbf{$e_{\textbf{smooth}} \downarrow$}  \\
\midrule
20 Hz (50 ms) & 87.45 &   63.74   & 11.90 \\
25 Hz (40 ms) & 84.57 &   62.80   & 9.49 \\
\midrule
\textbf{\textit{50 Hz (20 ms) (Ours)}} & \textbf{84.45} & \textbf{62.64} & \textbf{8.40} \\
\bottomrule
\end{tabular}
}
\label{tab:event_frequency_ablation}
\end{table}
}

\newcommand{\peractionresultstable}{
\begin{table*}[t!]
\caption{\textbf{Per-action quantitative results for the EE3D-R and EE3D-W datasets}.}
\centering
\resizebox{\textwidth}{!}{
\begin{tabular}{llc|cccccccccc|c}
\toprule
\textbf{Dataset} & \textbf{Method} & \textbf{Metric $\downarrow$} & \textbf{Walk} & \textbf{Crouch} & \textbf{Pushup} & \textbf{Boxing} & \textbf{Kick} & \textbf{Dance} & \textbf{Inter. w/ env.} & \textbf{Crawl} & \textbf{Sports} & \textbf{Jump} & \textbf{Avg.} \\
\midrule

\multirow{10}{*}{\textbf{EE3D-R}}
& \multirow{2}{*}{EgoPoseFormer \cite{yang2024egoposeformer}} 
  & MPJPE & 123.65 & 175.60 & 184.12 & 150.45 & 125.99 & 119.71 & 153.75 & 226.87 & 143.84 & 136.98 & 154.09 \\
&  & PA-MPJPE  & 78.70 & 107.19 & 109.20 & 102.66 & 91.67 & 86.74 & 87.95 & 111.85 & 96.71 & 99.60 & 97.22 \\
\cmidrule(lr){2-14}
& \multirow{2}{*}{EventEgo3D \cite{Millerdurai2024CVPR}} 
  & MPJPE & 74.75 & 144.26 & 109.58 & 141.19 & 104.33 & 89.98 & 103.15 & 118.30 & 108.23 & 105.77 & 109.95 \\
&  & PA-MPJPE  & 55.92 & 97.73 & 87.40 & 109.70 & 85.13 & 72.10 & 72.63 & 85.87 & 84.91 & 87.31 & 83.86 \\
\cmidrule(lr){2-14}
& \multirow{2}{*}{EventEgo3D++ \cite{Millerdurai2025IJCV}} 
  & MPJPE & 71.52 & 150.57 & 93.54 & 125.75 & 98.46 & 85.45 & 96.10 & 116.76 & 98.71 & 97.10 & 103.39 \\
&  & PA-MPJPE  & 52.35 & 99.96 & 71.12 & 94.82 & 79.93 & 65.97 & 68.60 & 81.29 & 76.16 & 76.56 & 76.67 \\
\cmidrule(lr){2-14}
& \multirow{2}{*}{\textbf{\textit{Ours (Causal)}}} 
  & MPJPE & \textbf{57.12} & \textbf{124.89} & \textbf{75.77} & \textbf{97.07} & \textbf{80.83} & \textbf{71.88} & \textbf{83.54} & \textbf{91.78} & \textbf{82.48} & \textbf{77.78} & \textbf{84.31} \\
&  & PA-MPJPE  & \textbf{39.49} & \textbf{86.09} & \textbf{57.63} & \textbf{74.61} & \textbf{62.25} & \textbf{52.84} & \textbf{59.37} & \textbf{65.66} & \textbf{63.02} & \textbf{60.52} & \textbf{62.14} \\
\cmidrule(lr){2-14}
& \multirow{2}{*}{\textbf{\textit{Ours (Non-Causal)}}} 
  & MPJPE & \textbf{52.79} & \textbf{121.08} & \textbf{73.73} & \textbf{94.06} & \textbf{76.49} & \textbf{68.81} & \textbf{80.77} & \textbf{88.42} & \textbf{79.60} & \textbf{76.01} & \textbf{81.17} \\
&  & PA-MPJPE  & \textbf{34.72} & \textbf{83.53} & \textbf{55.10} & \textbf{71.66} & \textbf{59.12} & \textbf{51.00} & \textbf{57.54} & \textbf{63.49} & \textbf{60.96} & \textbf{59.81} & \textbf{59.69} \\
\midrule

\multirow{10}{*}{\textbf{EE3D-W}} 
& \multirow{2}{*}{EgoPoseFormer \cite{yang2024egoposeformer}} 
  & MPJPE   & 176.96 & 251.27 & 263.46 & 215.26 & 180.34 & 171.24 & 220.04 & 324.68 & 205.83 & 188.64 & 220.40 \\
& & PA-MPJPE & 106.25 & 144.71 & 147.41 & 138.59 & 123.75 & 117.10 & 118.74 & 151.08 & 130.41 & 128.30 & 130.45 \\
\cmidrule(lr){2-14}
& \multirow{2}{*}{EventEgo3D \cite{Millerdurai2024CVPR}} 
  & MPJPE & 190.79 & 175.79 & 178.25 & 153.63 & 188.65 & 182.25 & 187.92 & 181.10 & 223.55 & 208.59 & 187.05 \\
& & PA-MPJPE & 105.58 & 107.45 & 112.51 & 77.66 & 101.43 & 108.35 & 101.95 & 98.08 & 122.16 & 112.70 & 104.78 \\
\cmidrule(lr){2-14}
& \multirow{2}{*}{EventEgo3D++ \cite{Millerdurai2025IJCV}} 
  & MPJPE & 166.32 & 155.40 & 166.60 & 141.25 & 157.74 & 159.02 & 163.52 & 149.61 & 206.15 & 179.59 & 164.52 \\
& & PA-MPJPE & 92.97 & 97.60 & 106.26 & 65.18 & 85.14 & 100.38 & 91.47 & 84.32 & 113.47 & 104.90 & 94.16 \\
\cmidrule(lr){2-14}
& \multirow{2}{*}{\textbf{\textit{Ours (Causal)}}} 
  & MPJPE & \textbf{144.64} & \textbf{138.75} & \textbf{148.76} & \textbf{131.62} & \textbf{146.04} & \textbf{142.56} & \textbf{149.34} & \textbf{138.67} & \textbf{192.47} & \textbf{168.50} & \textbf{150.13} \\
&  & PA-MPJPE  & \textbf{90.39} & \textbf{92.40} & \textbf{98.06} & \textbf{64.29} & \textbf{78.50} & \textbf{83.84} & \textbf{80.49} & \textbf{81.06} & \textbf{112.05} & \textbf{101.02} & \textbf{88.21} \\
\cmidrule(lr){2-14}
& \multirow{2}{*}{\textbf{\textit{Ours (Non-Causal)}}} 
  & MPJPE & \textbf{141.57} & \textbf{139.74} & \textbf{146.60} & \textbf{129.55} & \textbf{140.42} & \textbf{138.85} & \textbf{145.99} & \textbf{135.83} & \textbf{189.58} & \textbf{164.45} & \textbf{147.25} \\
&  & PA-MPJPE  & \textbf{88.24} & \textbf{91.40} & \textbf{96.60} & \textbf{61.15} & \textbf{74.56} & \textbf{80.73} & \textbf{77.49} & \textbf{78.10} & \textbf{109.65} & \textbf{98.21} & \textbf{85.61} \\
\bottomrule
\end{tabular}
}
\label{tab:per_action_results}
\end{table*}
}

\newcommand{\occlusiononlyresultstable}{
\begin{table*}[t]
\caption{\textbf{Quantitative results for occlusion-only end-effector joints for the EE3D-R and EE3D-W datasets}.} 
\centering
\resizebox{\textwidth}{!}{
\begin{tabular}{llc|cccccc}
\toprule
\textbf{Dataset} & \textbf{Method} & \textbf{Metric $\downarrow$} & \textbf{Elbow} & \textbf{Wrist} & \textbf{Knee} & \textbf{Ankle} & \textbf{Foot} & \textbf{Avg.} \\
\midrule

\multirow{8}{*}{\textbf{EE3D-R}}
& \multirow{2}{*}{EgoPoseFormer \cite{yang2024egoposeformer}} 
  & MPJPE     & 47.85  & 28.34  & 100.11 & 204.64 & 169.94 & 110.18 \\
& & PA-MPJPE  & 47.06  & 30.50  & 51.82  & 83.61  & 65.20  & 55.64 \\
\cmidrule(lr){2-9}
& \multirow{2}{*}{EventEgo3D \cite{Millerdurai2024CVPR}} 
  & MPJPE     & 37.46  & 21.97  & 79.20  & 156.93 & 138.95 & 86.90 \\
& & PA-MPJPE  & 40.41  & 25.78  & 48.58  & 78.27  & 61.63  & 50.93 \\
\cmidrule(lr){2-9}
& \multirow{2}{*}{EventEgo3D++ \cite{Millerdurai2025IJCV}} 
  & MPJPE     & 34.41  & 20.87  & 81.73  & 158.51 & 146.62 & 88.43 \\
& & PA-MPJPE  & 37.89  & 24.20  & 48.48  & 77.18  & 59.90  & 49.53 \\
\cmidrule(lr){2-9}
& \multirow{2}{*}{\textbf{\textit{Ours (Causal)}}}
  & MPJPE     & \textbf{29.91} & \textbf{16.12} & \textbf{60.53} & \textbf{120.54} & \textbf{108.37} & \textbf{67.49} \\
& & PA-MPJPE  & \textbf{31.68} & \textbf{19.02} & \textbf{41.08}  & \textbf{63.02} & \textbf{52.37}  & \textbf{41.85} \\
& \multirow{2}{*}{\textbf{\textit{Ours (Non-Causal)}}}
  & MPJPE  & \textbf{28.84} & \textbf{15.54} & \textbf{57.97} & \textbf{115.94} & \textbf{105.18} & \textbf{64.69} \\
& & PA-MPJPE  & \textbf{30.29} & \textbf{18.21}  & \textbf{39.09}  & \textbf{61.63} & \textbf{51.28} & \textbf{40.10} \\
\midrule

\multirow{8}{*}{\textbf{EE3D-W}} 
& \multirow{2}{*}{EgoPoseFormer \cite{yang2024egoposeformer}} & MPJPE     & 140.34 & 173.93 & 157.29 & 177.07 & 181.12  & 173.01 \\
& & PA-MPJPE & 104.34 & 119.89 & 102.39 & 124.28 & 132.86 & 113.67 \\
\cmidrule(lr){2-9}
& \multirow{2}{*}{EventEgo3D \cite{Millerdurai2024CVPR}} & MPJPE & 64.56  & 100.27 & 194.68 & 264.90 & 174.81 & 159.04 \\
& & PA-MPJPE & 61.44 & 101.60 & 76.81 & 105.15 & 79.18 & 84.84 \\
\cmidrule(lr){2-9}
& \multirow{2}{*}{EventEgo3D++ \cite{Millerdurai2025IJCV}} & MPJPE & 61.73  & 84.95 & 177.79 & 242.70 & 167.94 & 147.42 \\
& & PA-MPJPE & 57.28 & 87.80 & 74.33  & 95.53 & 73.24  & 77.64 \\
\cmidrule(lr){2-9}
& \multirow{2}{*}{\textbf{\textit{Ours (Causal)}}} & MPJPE & \textbf{54.04} & \textbf{75.69} & \textbf{162.53} & \textbf{229.09} & \textbf{157.84} & \textbf{135.84} \\
& & PA-MPJPE & \textbf{52.91} & \textbf{80.24}  & \textbf{70.06} & \textbf{92.96} & \textbf{69.78} & \textbf{73.19} \\
& \multirow{2}{*}{\textbf{\textit{Ours (Non-Causal)}}} & MPJPE & \textbf{54.72} & \textbf{74.84} & \textbf{160.40} & \textbf{225.28} & \textbf{156.34} & \textbf{134.72} \\
& & PA-MPJPE & \textbf{52.57} & \textbf{78.74}  & \textbf{68.90} & \textbf{92.11} & \textbf{69.14} & \textbf{72.29} \\

\bottomrule
\end{tabular}
}

\label{tab:occlusion_only_table}
\end{table*}
}

\newcommand{\perjointsresultstable}{
\begin{table*}[t!]
\caption{\textbf{Per-joint quantitative comparison for EE3D-R and EE3D-W datasets.}}
\centering
\setlength{\tabcolsep}{4pt}
\renewcommand{\arraystretch}{1.05}
\resizebox{\textwidth}{!}{%
\begin{tabular}{lll|*{9}{c}|c}
\toprule
\textbf{Dataset} & \textbf{Method} & \textbf{Metric $\downarrow$} &
Head & Neck & Shoulder & Elbow & Wrist & Hip & Knee & Ankle & Foot &
\textbf{Avg.} \\
\midrule

% ---------------- EE3D-R ----------------
\multirow{8}{*}{\textbf{EE3D-R}}
& EgoPoseFormer~\cite{yang2024egoposeformer} & MPJPE
  & 140.3 & 173.9 & 157.3 & 177.1 & 181.1 & 212.6 & 169.8 & 144.8 & 207.6 & 173.00 \\
&                                           & PA-MPJPE
  & 104.3 & 119.9 & 102.4 & 124.3 & 132.9 & 111.9 & 111.9 &  88.9 & 120.2 & 113.70 \\
\cmidrule(lr){2-13}
& EventEgo3D~\cite{Millerdurai2024CVPR}     & MPJPE
  & 23.49 & 29.87 & 40.20 & 91.67 & 153.17 & 69.99 & 120.96 & 179.02 & 201.42 & 110.39 \\
&                                           & PA-MPJPE
  & 49.27 & 41.63 & 44.48 & 79.58 & 133.53 & 55.66 & 83.45 & 108.78 & 125.26 & 84.52 \\
\cmidrule(lr){2-13}
& EventEgo3D++~\cite{Millerdurai2025IJCV}   & MPJPE
  & 22.66 & 29.36 & 40.03 & 77.86 & 125.41 & 63.02 & 119.86 & 177.03 & 196.58 & 103.29 \\
&                                           & PA-MPJPE
  & 43.01 & 36.52 & 40.54 & 71.09 & 112.89 & 48.17 & 79.38 & 105.13 & 119.58 & 77.07 \\
\cmidrule(lr){2-13}
& {\textbf{\textit{Ours (Causal)}}}         & MPJPE
  & \textbf{22.40} & \textbf{28.43} & \textbf{39.12} & \textbf{68.75} & \textbf{102.01} &
     \textbf{60.35} & \textbf{92.31} & \textbf{135.29} & \textbf{152.37} & \textbf{84.45} \\
&                                           & PA-MPJPE
  & \textbf{34.79} & \textbf{29.57} & \textbf{36.70} & \textbf{59.08} & \textbf{85.87} &
     \textbf{44.28} & \textbf{64.16} & \textbf{82.27} & \textbf{96.64} & \textbf{62.65} \\
& {\textbf{\textit{Ours (Non-Causal)}}}     & MPJPE
  & \textbf{22.06} & \textbf{28.44} & \textbf{37.95} & \textbf{66.43} & \textbf{98.21}  &
     \textbf{57.79} & \textbf{87.87} & \textbf{130.06} & \textbf{147.01} & \textbf{81.32} \\
&                                           & PA-MPJPE
  & \textbf{33.41} & \textbf{28.57} & \textbf{35.16} & \textbf{56.71} & \textbf{82.20} &
     \textbf{41.57} & \textbf{61.19} & \textbf{79.94} & \textbf{93.95} & \textbf{60.21} \\
\midrule

% ---------------- EE3D-W ----------------
\multirow{8}{*}{\textbf{EE3D-W}}
& EgoPoseFormer~\cite{yang2024egoposeformer} & MPJPE
  & 200.10 & 210.15 & 198.70 & 220.13 & 215.29 & 190.40 & 202.70 & 230.08 & 225.25 & 210.50 \\
&                                           & PA-MPJPE
  & 130.50 & 140.20 & 128.92 & 135.42 & 139.70 & 120.62 & 125.11 & 140.22 & 138.0 & 133.20 \\
\cmidrule(lr){2-13}
& EventEgo3D~\cite{Millerdurai2024CVPR}     & MPJPE
  & 46.54 & 61.64 & 82.57 & 145.81 & 228.19 & 161.96 & 239.96 & 315.86 & 335.56 & 195.50 \\
&                                           & PA-MPJPE
  & 69.16 & 55.87 & 65.48 & 102.11 & 184.56 & 76.49 & 94.26 & 134.92 & 145.26 & 108.20 \\
\cmidrule(lr){2-13}
& EventEgo3D++~\cite{Millerdurai2025IJCV}   & MPJPE
  & 44.87 & 56.67 & 74.02 & 127.95 & 185.59 & 136.72 & 215.69 & 285.08 & 303.66 & 172.43 \\
&                                           & PA-MPJPE
  & 59.46 & 49.49 & 59.01 & 90.39 & 157.96 & 64.35 & 93.59 & 128.17 & 139.41 & 98.42 \\
\cmidrule(lr){2-13}
& {\textbf{\textit{Ours (Causal)}}}         & MPJPE
  & \textbf{41.41} & \textbf{49.36} & \textbf{73.62} & \textbf{114.31} & \textbf{166.56} &
    \textbf{125.08} & \textbf{193.57} & \textbf{267.04} & \textbf{285.31} & \textbf{158.86} \\
&                                           & PA-MPJPE
  & \textbf{56.83} & \textbf{46.48} & \textbf{58.72} & \textbf{88.40} & \textbf{144.05} &
     \textbf{59.57} & \textbf{87.36} & \textbf{123.67} & \textbf{134.30} & \textbf{93.46} \\
& {\textbf{\textit{Ours (Non-Causal)}}}     & MPJPE
  & \textbf{43.74} & \textbf{51.32} & \textbf{73.01} & \textbf{112.67} & \textbf{163.55} &
    \textbf{122.79} & \textbf{189.22} & \textbf{259.93} & \textbf{277.88} & \textbf{155.82} \\
&                                           & PA-MPJPE
  & \textbf{55.95} & \textbf{45.28} & \textbf{57.46} & \textbf{85.13} & \textbf{138.42} &
     \textbf{58.55} & \textbf{85.02} & \textbf{121.06} & \textbf{130.62} & \textbf{90.86} \\
\midrule

\end{tabular}%
}
\label{tab:per_joint_results}
\end{table*}
}

\newcommand{\methodfigure}{
\begin{figure*}
    \centering    \includegraphics[trim={0cm 2.8cm 0cm 2.8cm},clip,width=\textwidth]{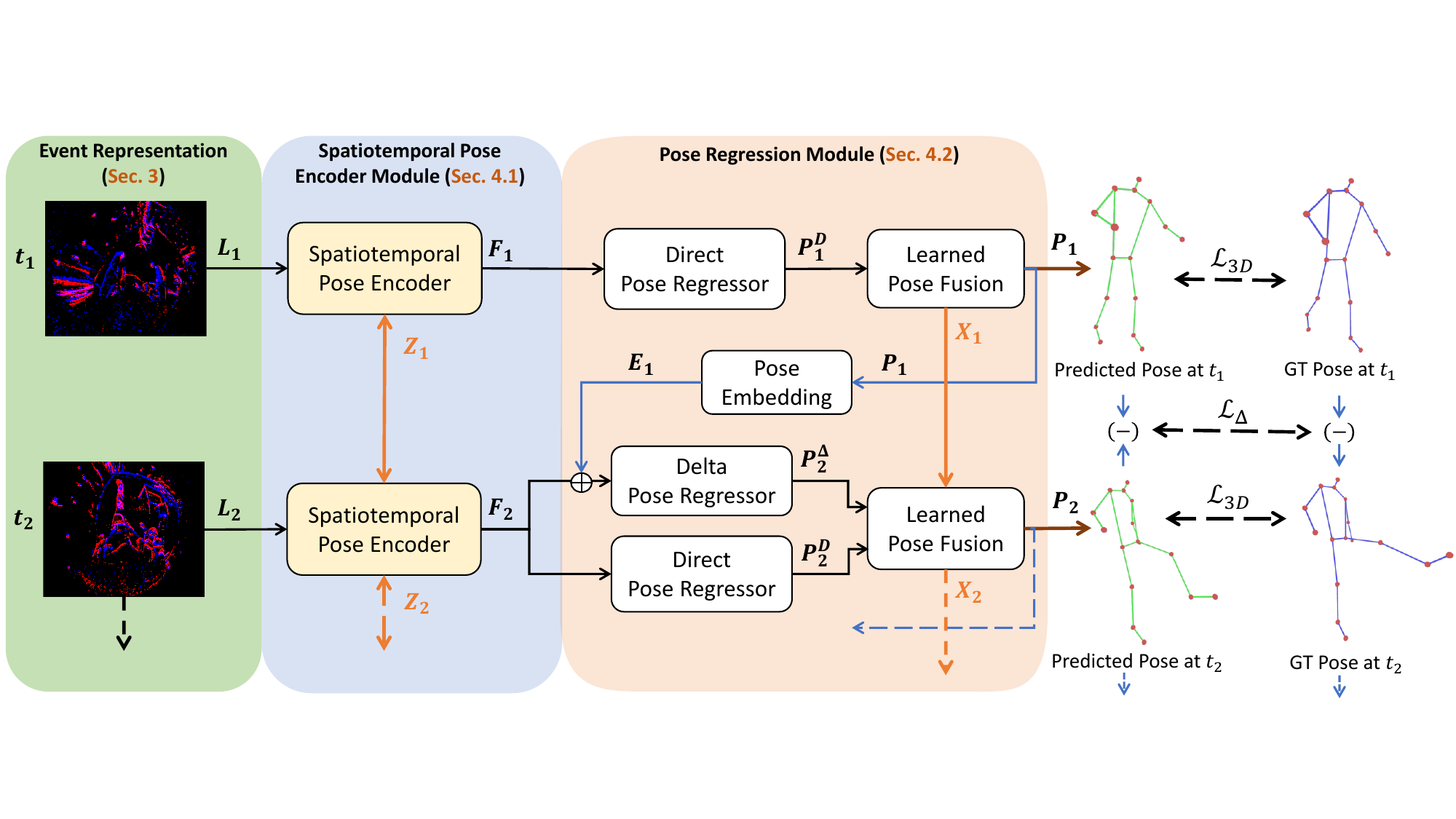}
    \vspace{-20pt}
    \caption{\textbf{Overview of the proposed \textit{E-3DPSM} approach for monocular egocentric 3D human pose estimation.} Incoming raw events $e$ are converted into LNES frames $\mathbf{L}_t$ and processed by the Spatiotemporal Pose Encoder Module (SPEM, \cref{sec:spem}), 
    as depicted in~\cref{fig:SPEM}. 
    The output of SPEM $\mathbf{F}_t$ is passed to the Pose Regression Module (PRM, \cref{sec:prm}), which estimates both the direct 3D poses $\mathbf{P}^{D}_t$ and the 3D delta poses $\mathbf{P}^{\Delta}_t$ between consecutive LNES frames. These delta poses naturally correspond to the observed events (i.e., changes in the 2D LNES space). 
     Finally, the 3D pose deltas are fused with the direct 3D pose using a learned pose fusion mechanism (a neural Kalman-style filter) to obtain the final 3D poses $\mathbf{P}_t$.
     The predicted 3D pose $\mathbf{P}_t$ is supervised using absolute ground-truth 3D poses ($\mathbf{\mathcal{L}_{\text{3D}}}$) and ground-truth 3D pose differences ($\mathcal{L}_{\Delta}$), respectively.
    At each timestep $t$, SPEM updates and propagates its latent state $\mathbf{Z}_t$ bidirectionally (the ``$\longleftrightarrow$'' symbol), whereas PRM propagates its fused pose state $\mathbf{X}_t$ causally (the ``$\longrightarrow$'' symbol) to the next timestep.
    During training, SPEM operates bidirectionally (non-causal) 
    to exploit the full temporal context, while at inference, it can run in either causal (real-time) or non-causal mode. 
    For visualisation purposes, the differences between 
    $\textbf{P}_1$
    and $\textbf{P}_2$ and the corresponding $\textbf{L}_1$
    and $\textbf{L}_2$
    are exaggerated.  
    } 
    \vspace{-10pt}
    \label{fig:method_figure}
\end{figure*}
}

\newcommand{\mpjpequalitativefigure}{
\begin{figure*}
    \centering   \includegraphics[trim={6.8cm 5cm 3.5cm 3.2cm},clip,width=\textwidth]{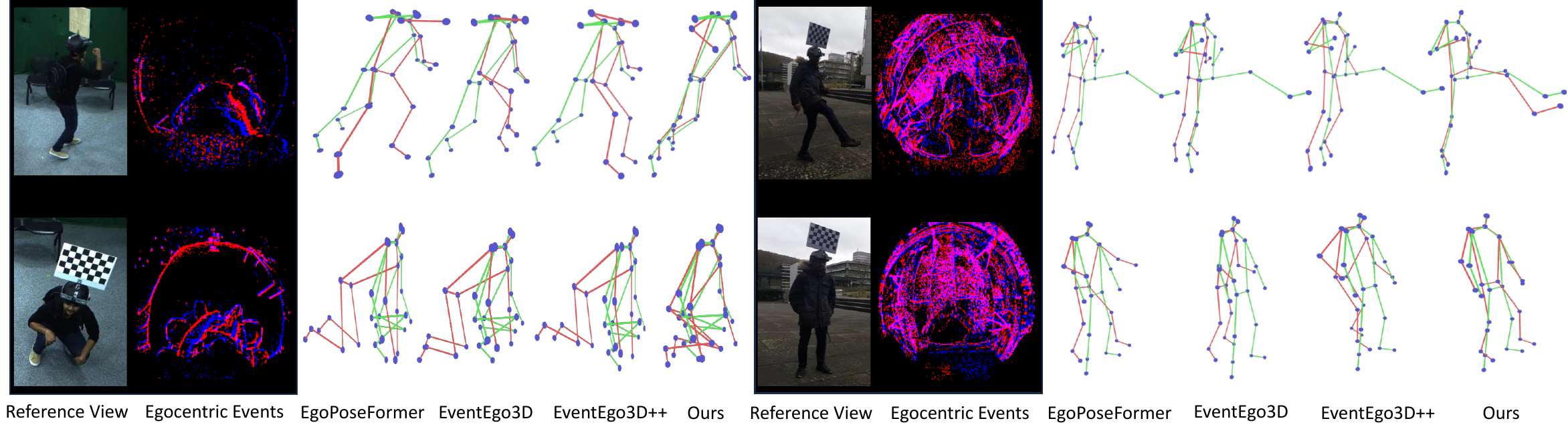}
    \vspace{-20pt}
    \caption{\textbf{Qualitative comparison of our method with prior approaches.} We compare against EgoPoseFormer~\cite{yang2024egoposeformer}, EventEgo3D~\cite{Millerdurai2024CVPR}, and EventEgo3D++~\cite{Millerdurai2025IJCV}. 
    \textbf{Left:} EE3D-R (real dataset). \textbf{Right:} EE3D-W (in-the-wild). \textbf{Red:} Predicted pose.  \textbf{Green:} Ground truth. 
    }
    \vspace{-15pt}
    \label{fig:qualitative_results}
\end{figure*}
}

\newcommand{\jitterqualitativefigure}{
\begin{figure}[t]
    \centering
    \includegraphics[width=\columnwidth]{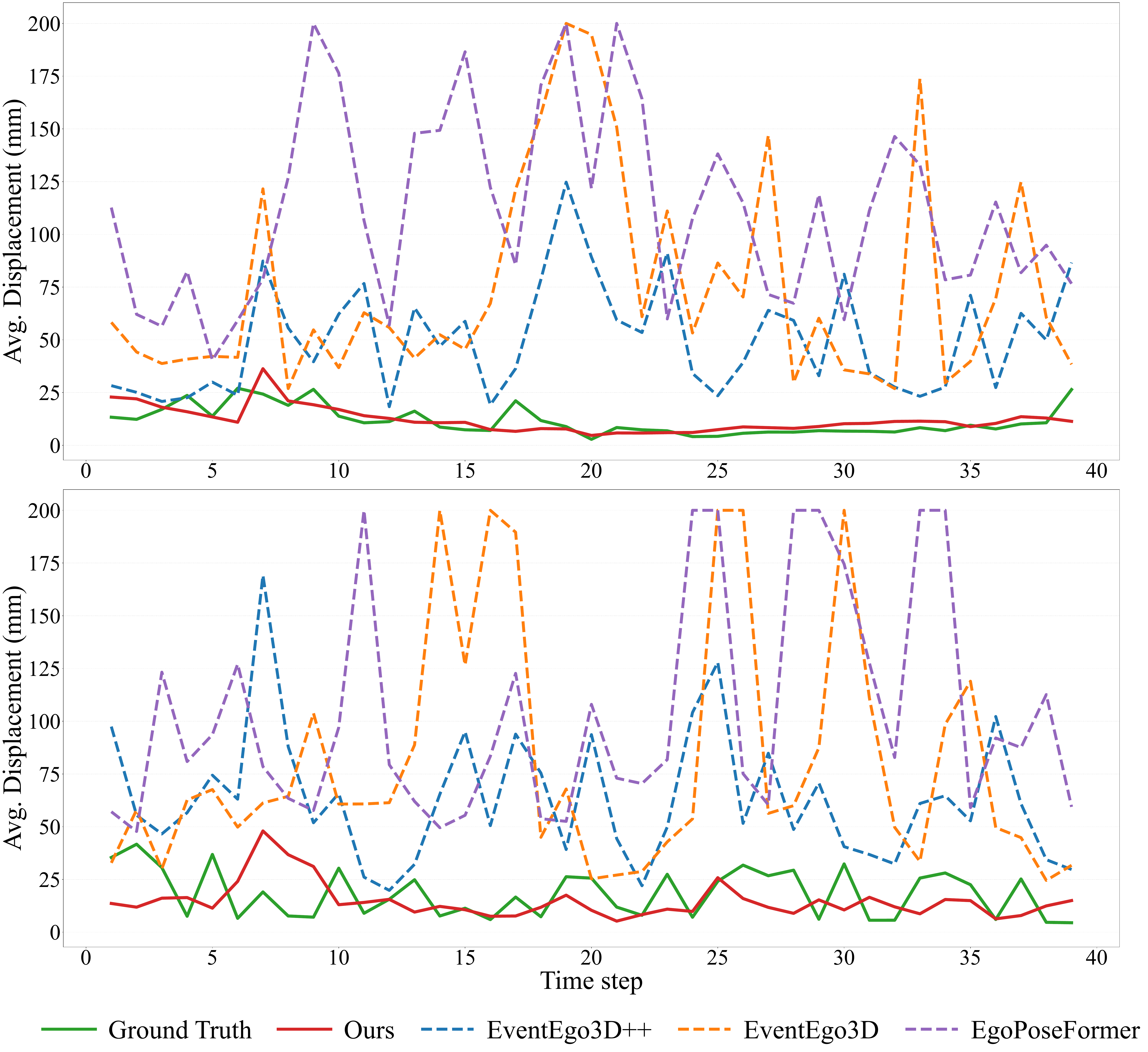}
    \vspace{-15pt}
    \caption{We plot the per-frame all-joint average displacement (Eq.~\eqref{eq:jitter_eqn}) for EE3D-R (top) and EE3D-W (bottom). 
    }
    \vspace{-15pt}
    \label{fig:qualitative_jitter_results}
\end{figure}
}

\newcommand{\encoderfigure}{
\begin{figure}[!t]
    \centering
    \raggedleft
    \includegraphics[width=\columnwidth]{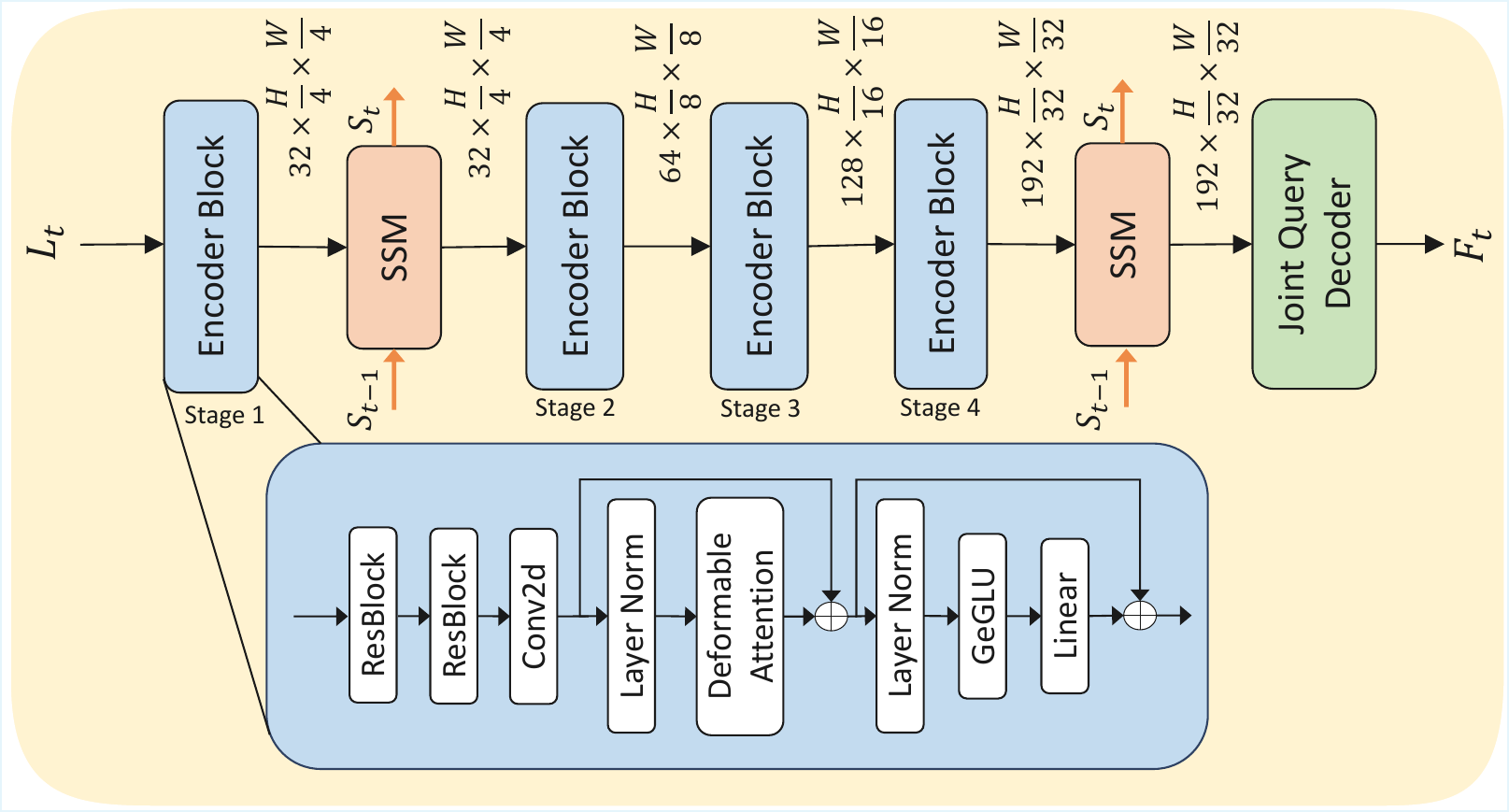}
    \vspace{-20pt}
    \caption{\textbf{Architecture of 
    SPEM},
    combining multi-stage convolutional encoding, SSM blocks, deformable attention, and a joint-query decoder for temporally-aware pose features.}
    \vspace{-15pt}
    \label{fig:SPEM}
\end{figure}
}

\newcommand{\fusionerrorfigure}{
\begin{figure}[t]
    \centering
    \raggedleft
    \includegraphics[width=\columnwidth]{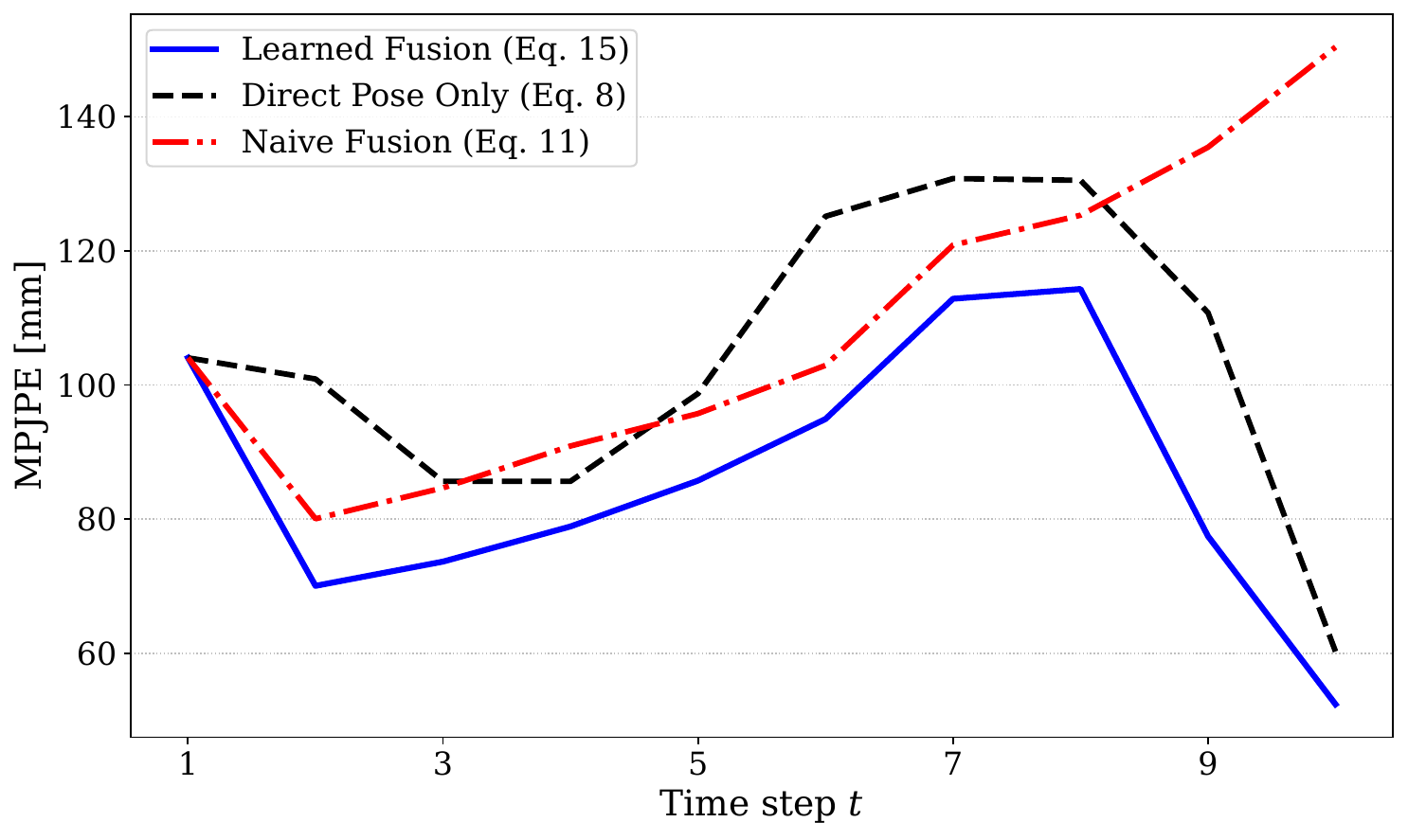}
    \caption{\textbf{Pose drift over time.} Comparison of learned fusion (Eq.~\eqref{eq:learned_fusion}), direct pose only (Eq.~\eqref{eq:abs_pose}), and naive fusion (Eq.~\eqref{eq:naive_fusion}) across temporal sequence length. Naive fusion leads to rapidly increasing drift, whereas our learned fusion effectively mitigates this drift, maintaining stable accuracy over time.}
    \label{fig:error_accumulation}
\end{figure}
}

\newcommand{\jitterperjointeedr}{
\begin{figure}[!t]
    \centering
    \raggedright % Ensures right alignment
    \includegraphics[width=\linewidth]{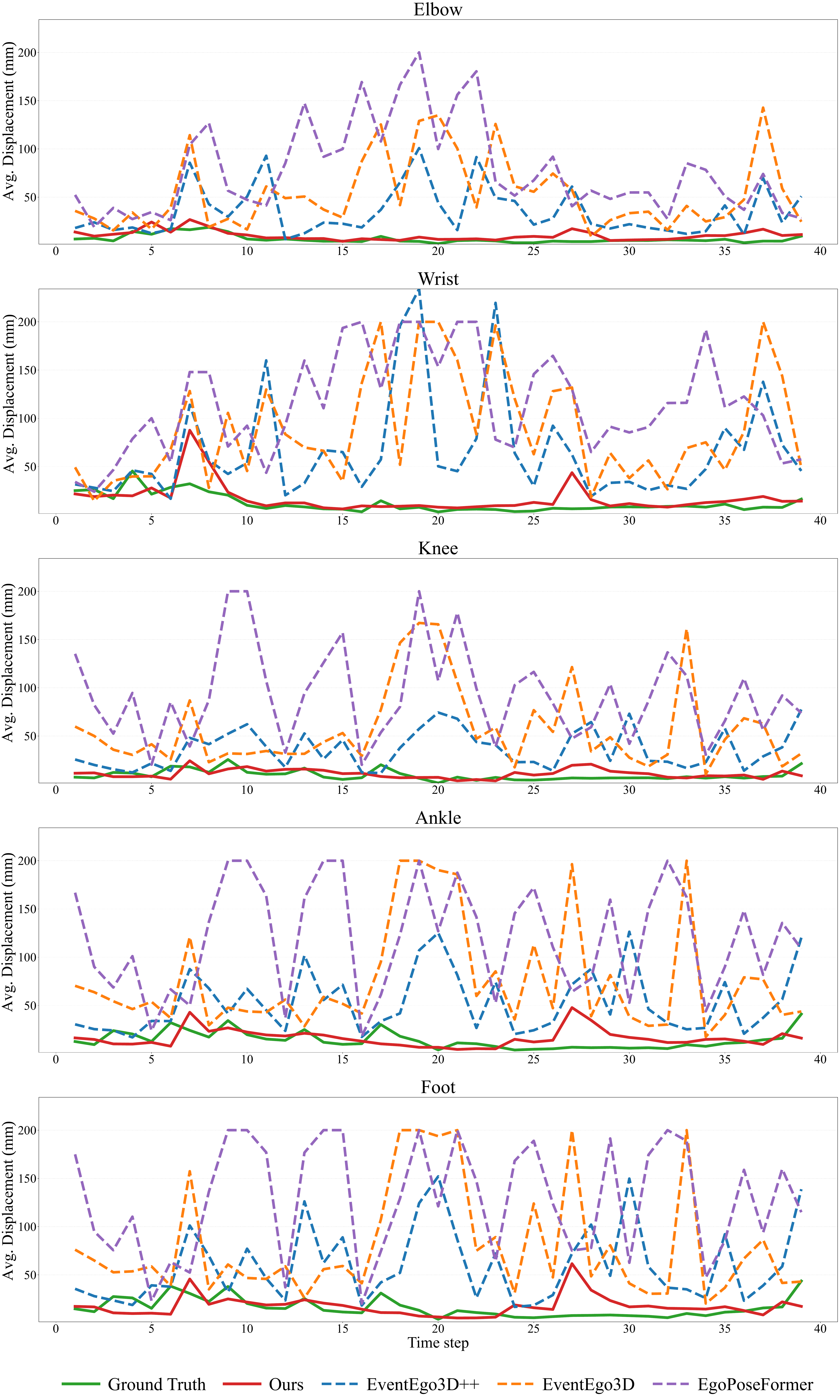}
    \caption{\textbf{Qualitative results showing jitter improvement on EE3D-R}}
    \label{fig:qualitative_ee3dr_joints_jitter_results}
\end{figure}
}

\newcommand{\jitterperjointeedw}{
\begin{figure}[!t]
    \centering
    \raggedright % Ensures right alignment
    \includegraphics[width=\linewidth]{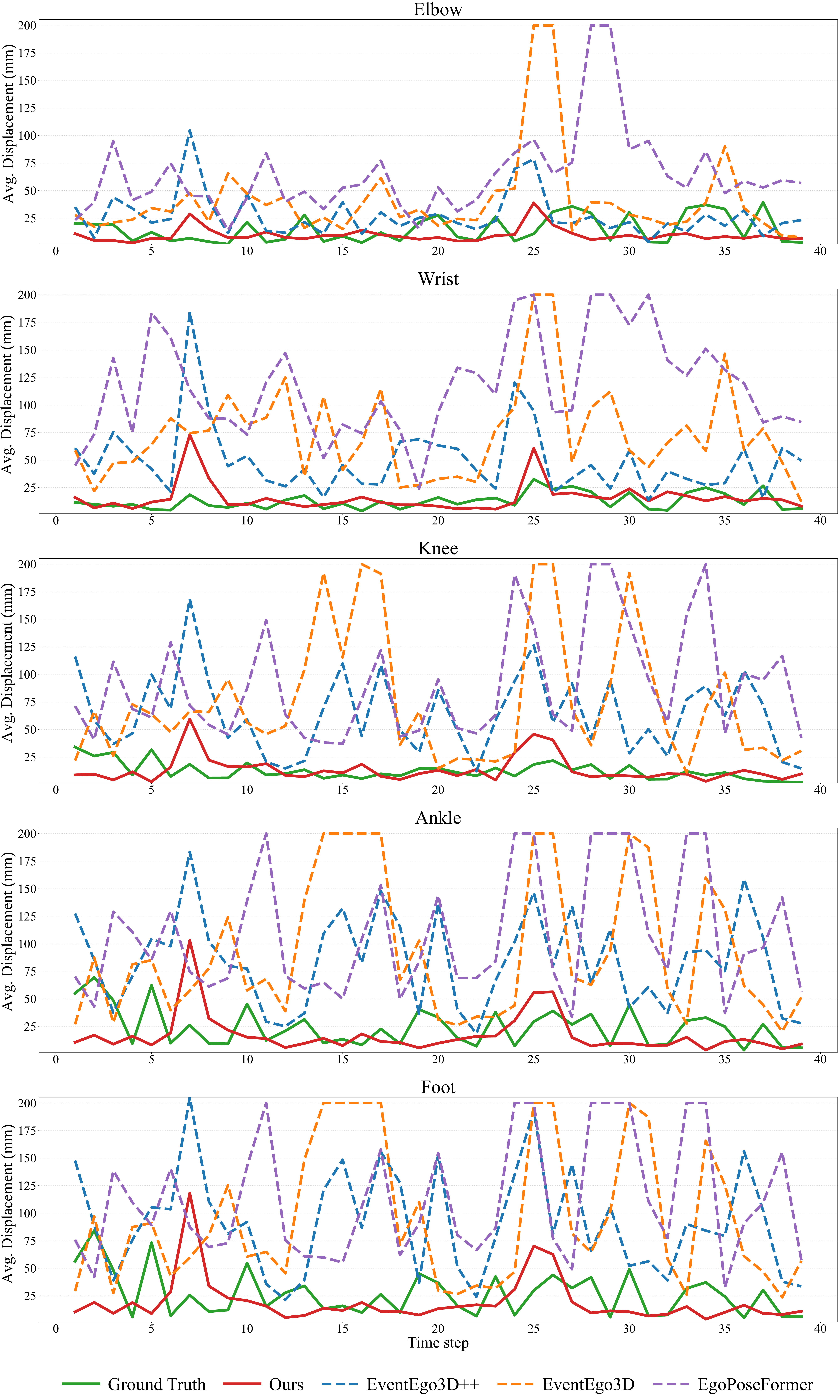}
    \caption{\textbf{Qualitative results showing jitter improvement on EE3D-W}}
    \label{fig:qualitative_ee3dw_joints_jitter_results}
\end{figure}
}

\newcommand{\jitterfigs}{
\begin{figure*}[t]
    \centering

    % Left figure: EE3D R
    \begin{minipage}[t]{0.48\linewidth}
        \centering
        \includegraphics[width=\linewidth]{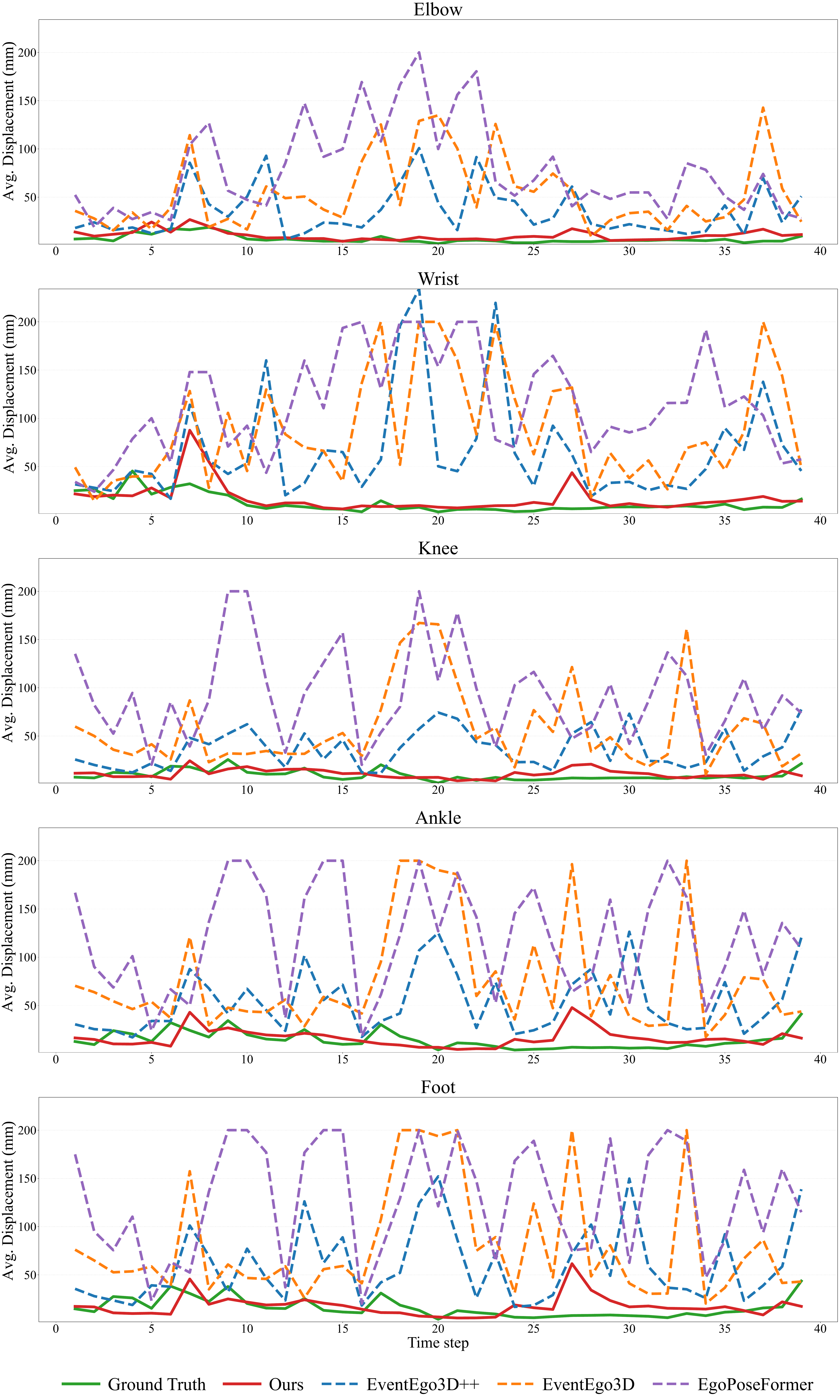}
        \captionof{figure}{The per-frame average end-effector joint displacements (Eq.~\eqref{eq:jitter_eqn}) for EE3D-R. Zoom recommended.}
        \label{fig:qualitative_ee3dr_joints_jitter_results}
    \end{minipage}
    \hfill
    % Right figure: EE3D W
    \begin{minipage}[t]{0.48\linewidth}
        \centering
        \includegraphics[width=\linewidth]{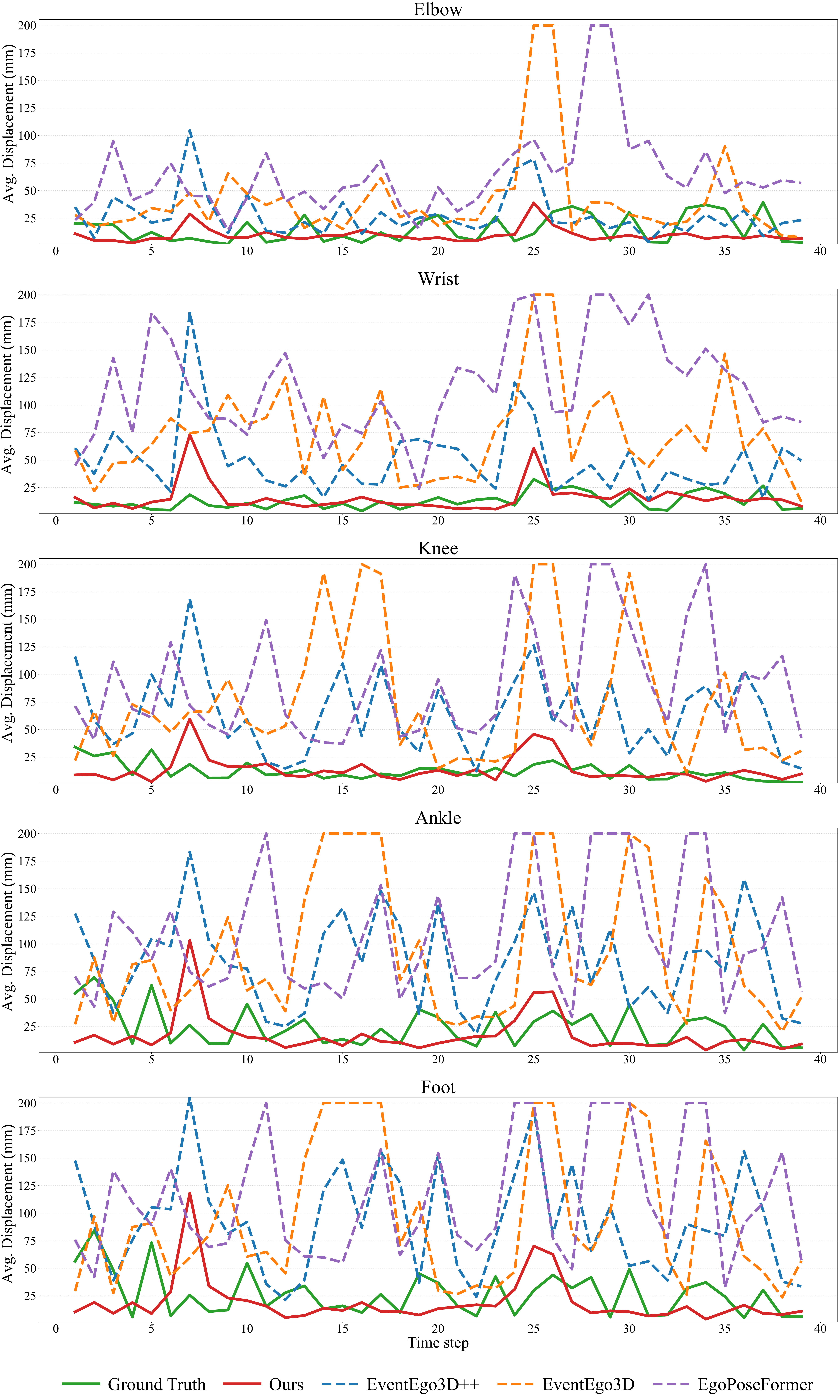}
        \captionof{figure}{The per-frame average end-effector joint displacements (Eq.~\eqref{eq:jitter_eqn}) for EE3D-W. Zoom recommended.}
        \label{fig:qualitative_ee3dw_joints_jitter_results}
    \end{minipage}

\end{figure*}
}

\newcommand{\qualitativewildfigure}{
\begin{figure*}[t]
    \centering
    \vspace{-10pt}
    \includegraphics[trim={1.5cm 2.5cm 1.5cm 0.2cm},clip,width=\textwidth]{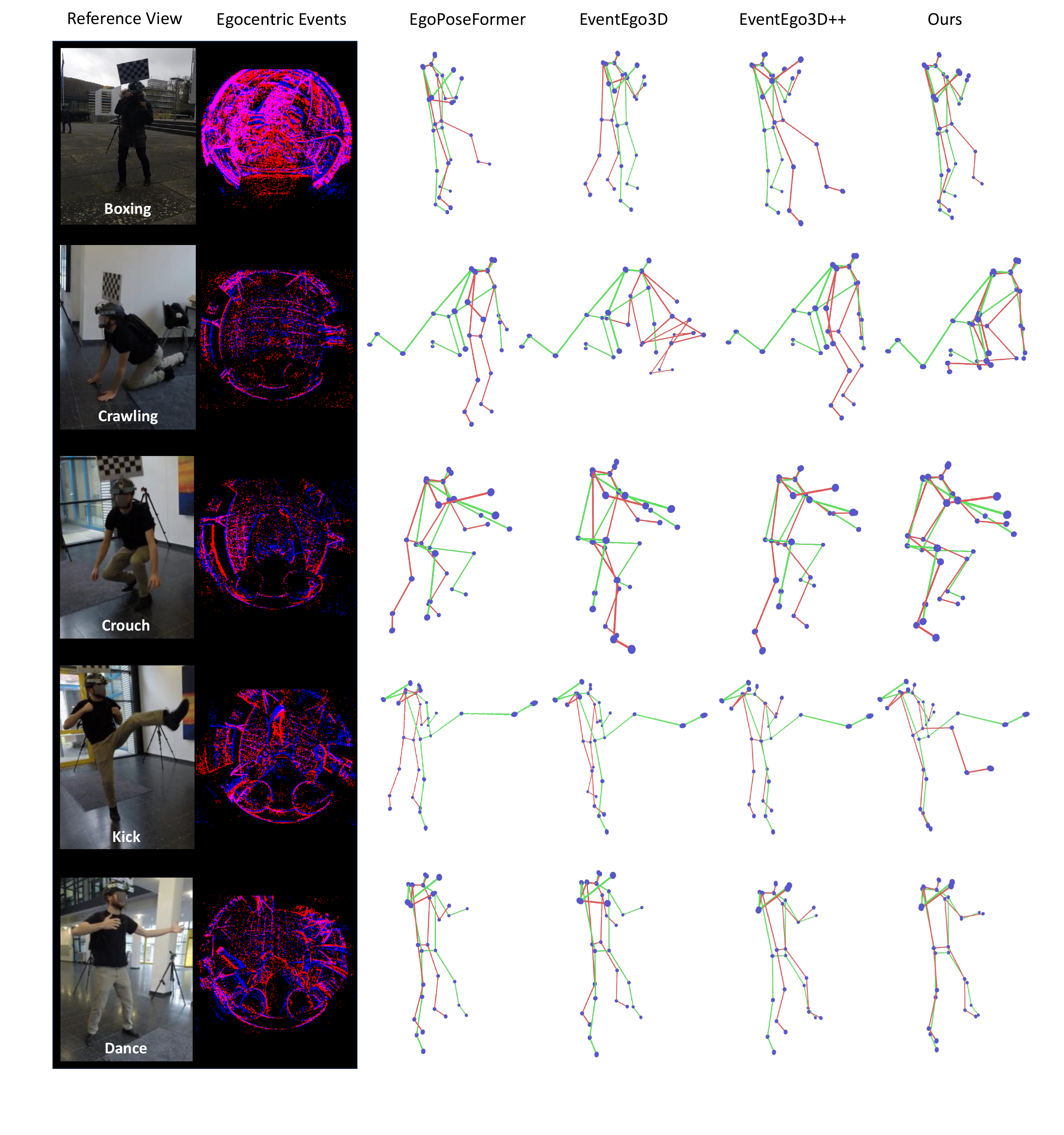}
    \vspace{-12pt}
    \caption{\textbf{Per-action qualitative comparison of our method with prior approaches on EE3D-W (challenging sequences).} We compare against EgoPoseFormer~\cite{yang2024egoposeformer}, EventEgo3D~\cite{Millerdurai2024CVPR}, and EventEgo3D++~\cite{Millerdurai2025IJCV}. \textbf{Red:} Predicted pose. \textbf{Green:} Ground truth.}
    \label{fig:qualitative_wild}
    \vspace{-10pt}
\end{figure*}
}

\newcommand{\qualitativerealfigure}{
\begin{figure*}[t]
    \centering
    \vspace{-10pt}
    \includegraphics[trim={1.5cm 2.5cm 2.3cm 0.2cm},clip,width=\textwidth]{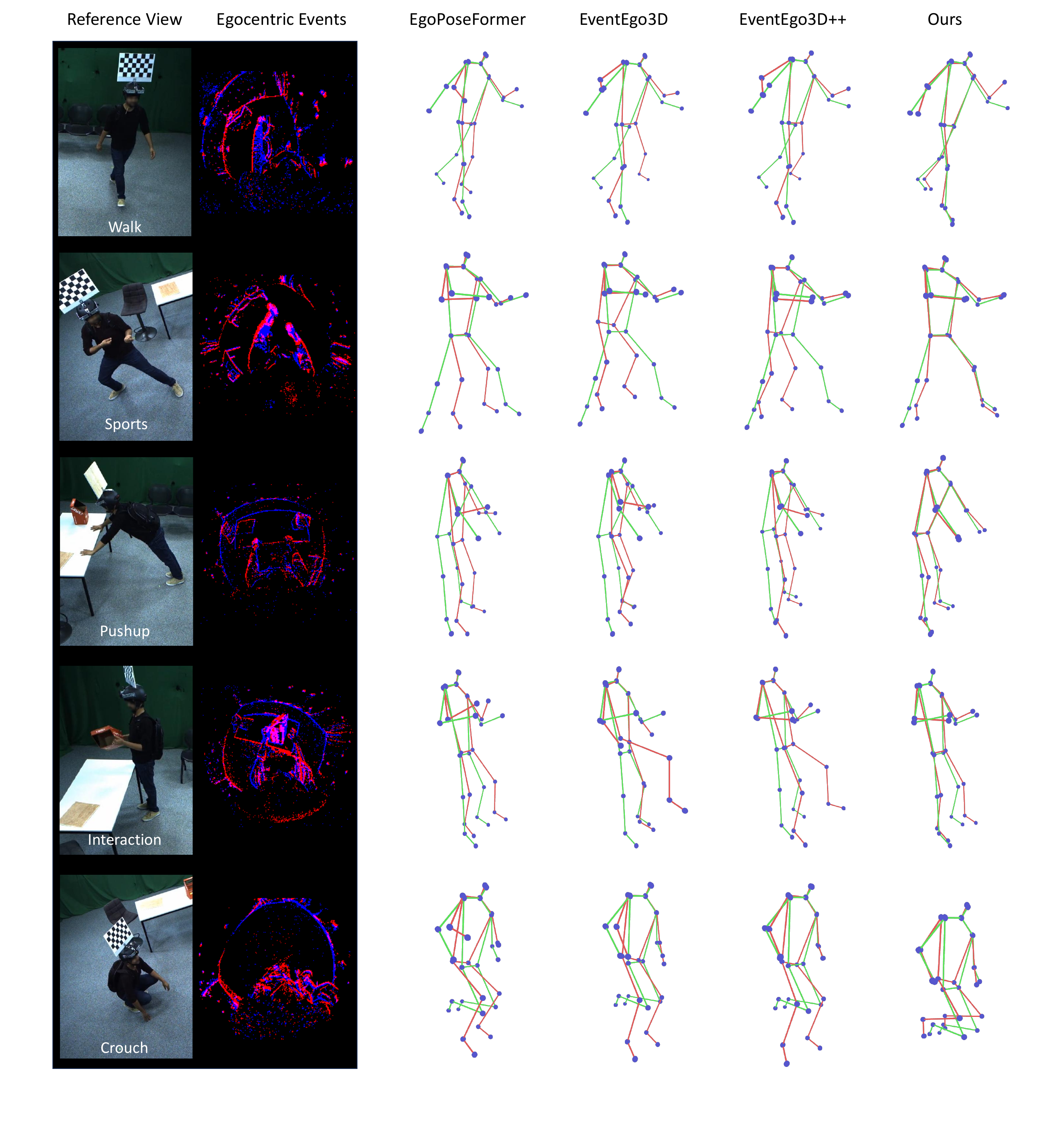}
    \vspace{-12pt}
    \caption{\textbf{Per-action qualitative comparison of our method with prior approaches on EE3D-R (walk and further challenging sequences).} We compare against EgoPoseFormer~\cite{yang2024egoposeformer}, EventEgo3D~\cite{Millerdurai2024CVPR}, and EventEgo3D++~\cite{Millerdurai2025IJCV}. \textbf{Red:} Predicted pose.  \textbf{Green:} Ground truth.}
    \label{fig:qualitative_real}
    \vspace{-10pt}
\end{figure*}
}

\newcommand{\hmdsetup}{
\begin{figure}[!t]
    \centering
    \includegraphics[trim={1cm 1cm 1.5cm 0cm},clip,width=\columnwidth]{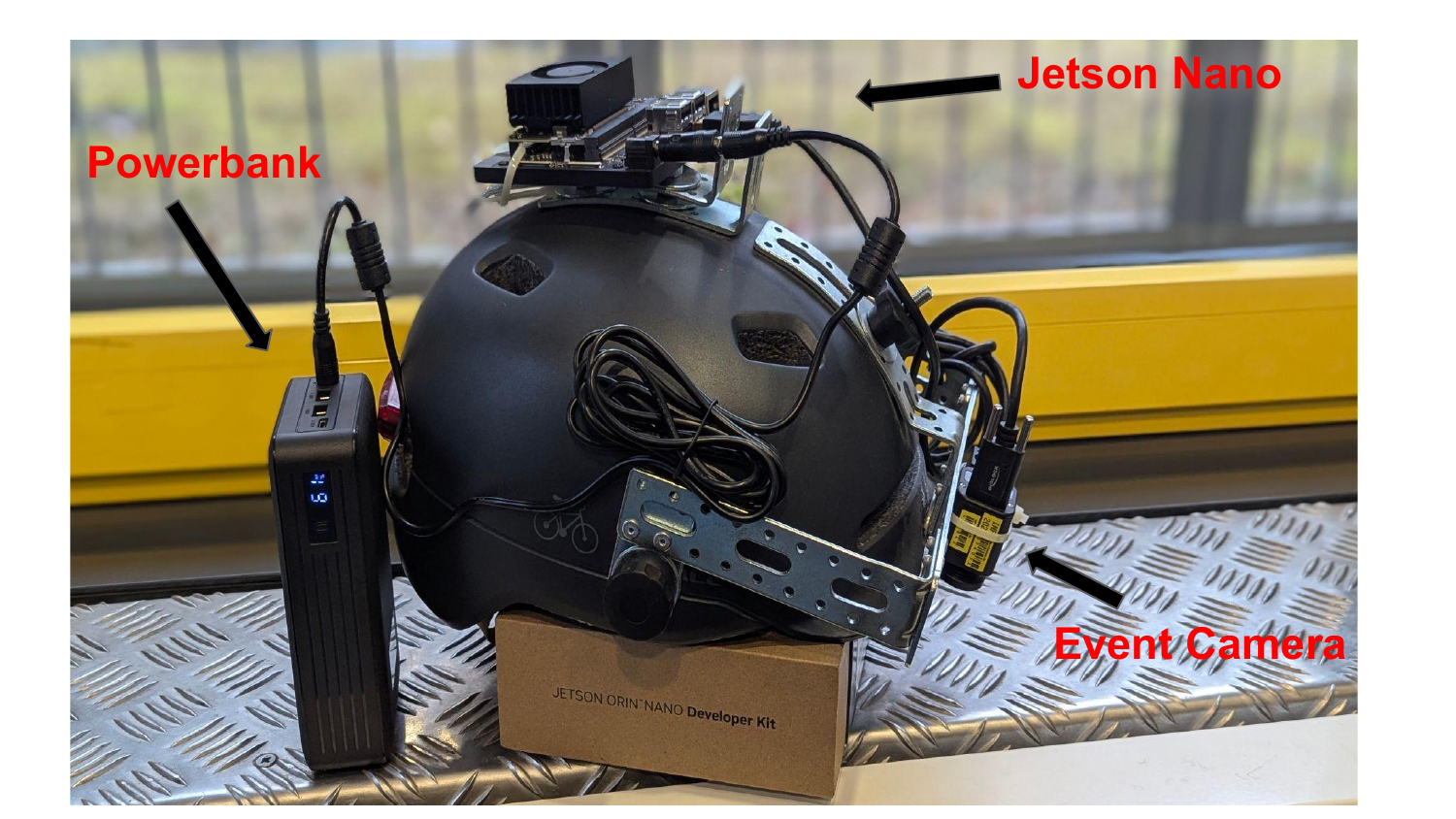}
    \caption{\textbf{Our head-mounted device setup.} The device uses a single fisheye egocentric event camera for input, NVIDIA Jetson Orin Nano for onboard processing, and a portable powerbank for standalone operation.}
    \label{fig:hmd_setup}
\end{figure}
}

\newcommand{\poganalysis}{
\begin{figure}[!t]
    \centering
    \includegraphics[trim={0cm 0.2cm 0cm 0cm},clip,width=\columnwidth]{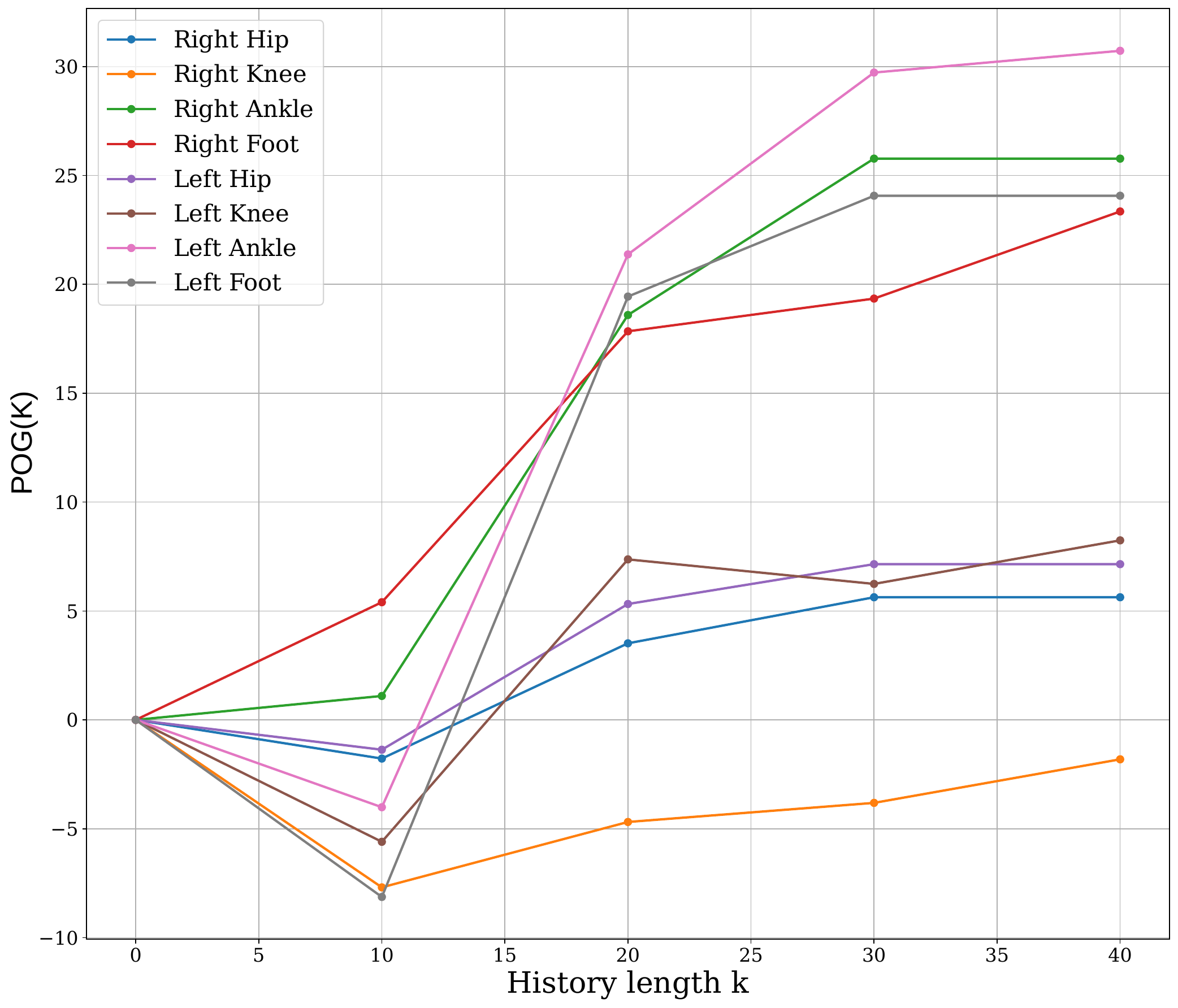}
    \caption{We plot the improvement in MPJPE obtained by increasing the duration of temporal history k, showing how a longer past context yields larger gains for occluded lower body joints.}
    \label{fig:pog_analysis}
\end{figure}
}

\newcommand{\limitationfigure}{
\begin{figure*}
    \centering   \includegraphics[trim={1.5cm 8cm 1.5cm 6cm},clip,width=\textwidth]{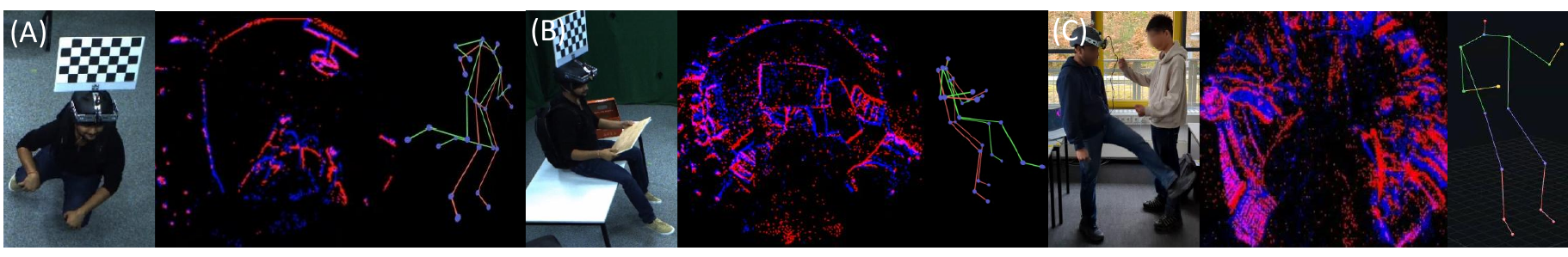}
    \vspace{-20pt}
    \caption{\textbf{Failure cases for different scenarios.} (A) Strong self-occlusion crawl action, (B) interaction with objects, (C) other humans in the FOV. External views are only for reference. \textbf{Red}: Predicted pose. \textbf{Green}: Ground truth. 
    C visualises our prediction only (no ground truth available). 
    \textit{Inputs to E-3DPSM are egocentric LNES frames.}}
    \vspace{-15pt}
    \label{fig:limitationfigure}
\end{figure*}
}

\newcommand{\viewerdemo}{
\begin{figure}[t]
    \centering
    \raggedleft
    \includegraphics[width=\columnwidth]{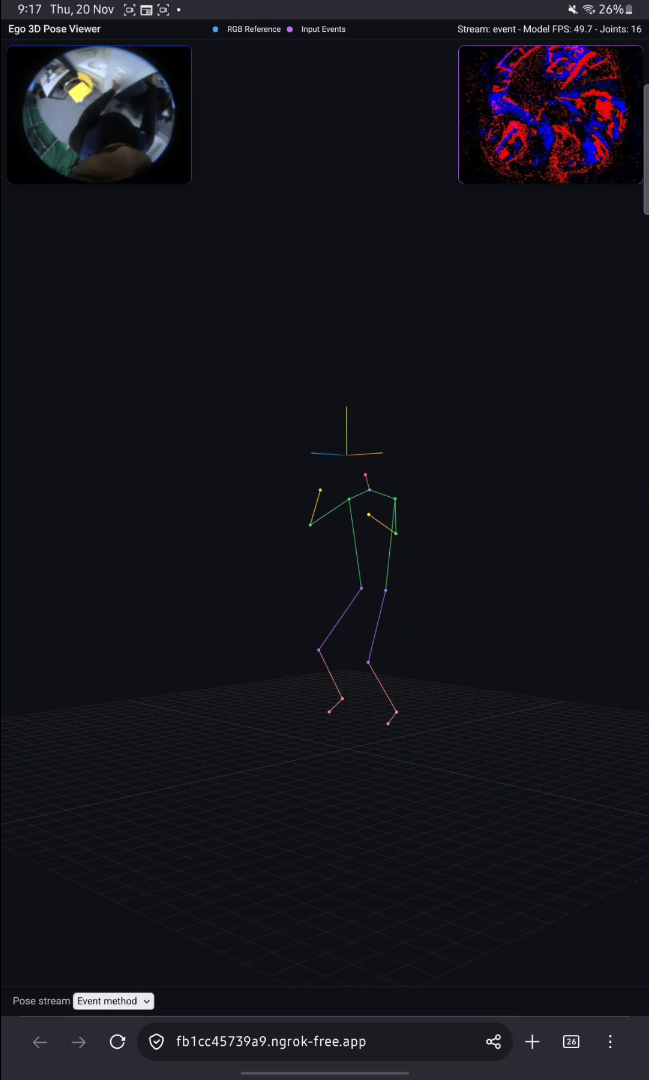}
    \caption{\textbf{Our real-time viewer.} Screenshot of our iPad-viewer showing the live event stream, reference RGB view, and the predicted 3D skeleton rendered in real time. Note that there is a transmission delay of 3--5 poses.}
    \label{fig:viewerdemo}
\end{figure}
}

\maketitle

\begin{abstract}
Event cameras offer multiple advantages in monocular egocentric 3D human pose estimation from head-mounted devices, such as millisecond temporal resolution, high dynamic range, and negligible motion blur. 
Existing methods effectively leverage these properties, but suffer from low 3D estimation accuracy, insufficient in many applications (e.g., immersive VR/AR). 
This is due to the design not being fully tailored towards event streams (e.g., their asynchronous and continuous nature), 
leading to high sensitivity to self-occlusions and temporal jitter in the estimates. 
This paper rethinks the setting and 
introduces \hbox{E-3DPSM}, an event-driven continuous pose state machine for event-based egocentric 3D human pose estimation.
\hbox{E-3DPSM} aligns continuous human motion with fine-grained event dynamics; it evolves latent states and predicts continuous changes in 3D joint positions associated with observed events, which are fused with direct 3D human pose predictions, leading to stable and drift-free final 3D pose reconstructions. 
\hbox{E-3DPSM} runs in real-time at $80$ Hz on a single workstation and sets a new state of the art in experiments on two benchmarks, improving accuracy 
by up to $19\%$
% over 10\% 
(MPJPE) and temporal stability by up to $2.7\times$. 
\textit{See our project page for the source code and trained models\footnote{\url{https://4dqv.mpi-inf.mpg.de/E-3DPSM/}}.}
% \textit{The source code and trained models are available on our project page\footnote{\url{https://4dqv.mpi-inf.mpg.de/E-3DPSM/}}.} 
% 
\end{abstract}

\begin{figure}[!t]
  \centering
  \includegraphics[trim={8cm 6.5cm 1.8cm 0.2cm},clip, width=\columnwidth]{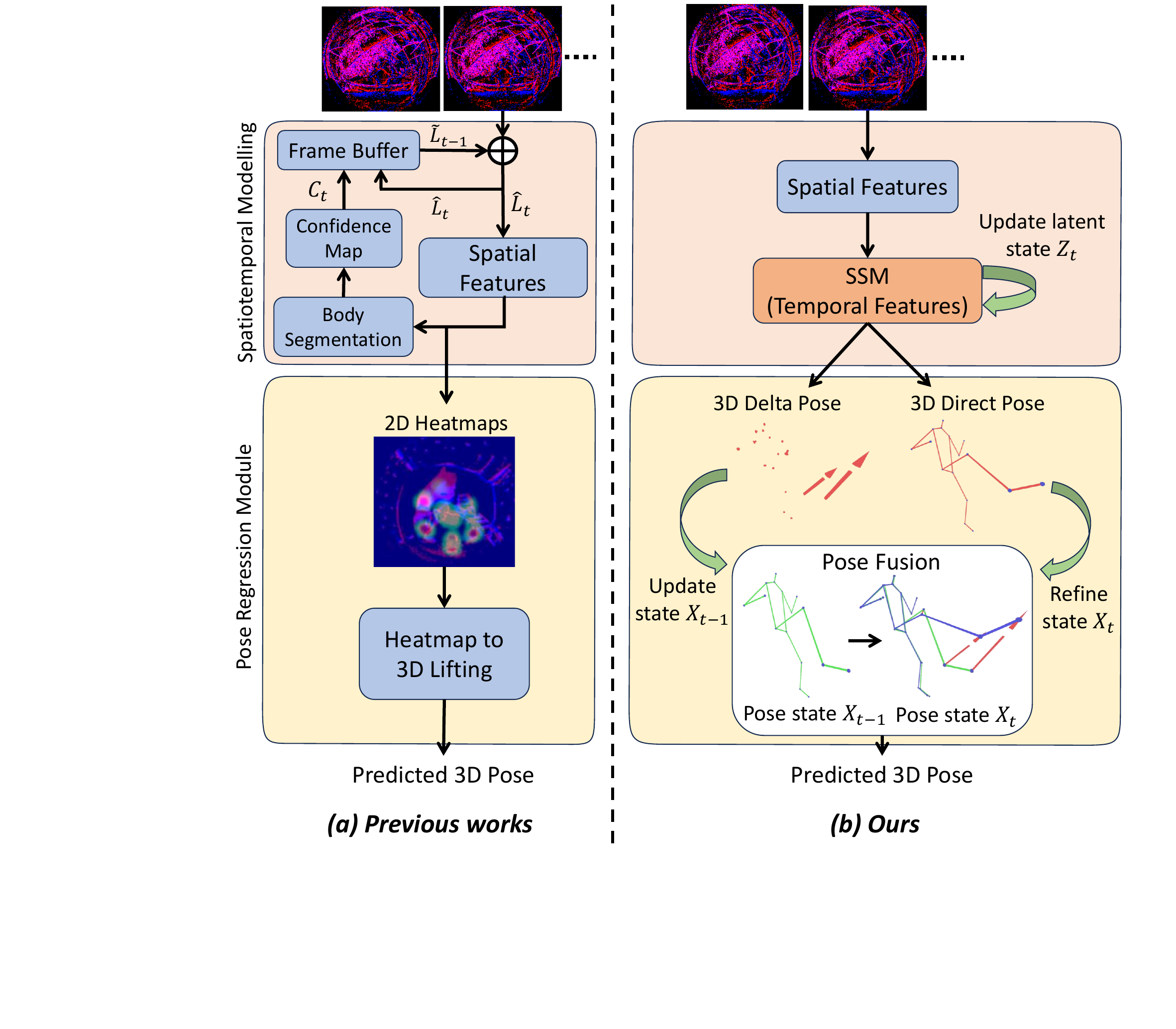}
  \vspace{-18pt}
  \caption{
  \textbf{Rethinking event-based egocentric 3D human pose estimation.} 
  (a) Previous methods \cite{Millerdurai2024CVPR, Millerdurai2025IJCV} capture temporal information only through a single previous event frame stored in the frame buffer 
  leading to jitter and drift. (b) Our E-3DPSM approach models motion as a continuous event-driven state evolution, fusing delta and direct 3D human pose updates, thereby achieving real-time and temporally stable 3D reconstruction and significantly outperforming prior approaches in the 3D accuracy.} 
  \vspace{-10pt}
  \label{fig:teaser}
\end{figure}

\section{Introduction} 
\label{sec:intro}

Egocentric 3D human pose estimation 
from head-mounted devices (HMDs) is a key capability for immersive VR/AR applications such as real-time avatar control, fitness tracking, telepresence, and hands-free interfaces. By capturing motion directly from the wearer’s perspective, it removes the need for external cameras and constrained capture environments. 
Yet, fast camera motion and frequent self-occlusions also add new algorithmic challenges. 

Recent RGB-based egocentric 3D pose estimation methods~\cite{yang2024egoposeformer,hakada2025egorear,hakada2024unrealego2,hakada2022unrealego,9711289,kang2024egotap,10.1145/3610548.3618147,liuegofish3d,Luo2021DynamicsRegulatedKP,10.1145/3580305.3599312,10.1145/2980179.2980235,9010983,10.1109/TPAMI.2020.3029700,Wang_2021_ICCV,wang2022estimating,wang2024egocentric,yuan2019ego,9665856} achieve accurate results in controlled, well-lit environments, but struggle under real-world conditions. Low light causes underexposure and sensor noise, rapid head motion leads to blur, and continuous streaming of high-resolution video imposes heavy bandwidth and power demands on wearable devices. 
As a viable alternative in challenging conditions, event cameras offer millisecond-level temporal resolution and high dynamic range. 
The recently introduced EventEgo3D \cite{Millerdurai2024CVPR, Millerdurai2025IJCV} 
is the first method for egocentric and real-time 3D human pose estimation from a single head-mounted event camera. 
Compared to third-person captures \cite{Wang_2025_ICCV}, humans remain centred and scale-consistent across input frames in egocentric views. 
These properties reduce background variability and make egocentric event streams particularly suitable for continuous
3D reconstruction of fast human motions. 
While EventEgo3D can capture 3D human poses at high temporal resolutions, its overall accuracy is considered low in many practical scenarios.  
We believe the reasons are in its architecture (\cref{fig:teaser}-(a)): First, EventEgo3D predicts direct 3D poses, models temporal information using only the previous event frame stored in the frame buffer through short-term feature propagation across event frames and, hence, 
does not fully exploit the asynchronous, continuous, and change-driven nature of event data. 
This often leads to jitter, drift, and substantial 3D errors under self-occlusions. 
Moreover, the reliance on 2D heatmaps introduces quantisation errors ~\cite{Zhang_2020_CVPR}, and  
segmentation masks that need to be predicted at test time 
can serve as an additional source of inaccuracies. 

This paper fundamentally rethinks egocentric event-based 3D human pose estimation with the overarching goal of improving the 3D estimation accuracy. 
Our key insights are threefold:
1) Since events inherently encode changes in the 2D observation space, they should correspond to and induce changes in the 3D space. 
The changes in the relative 3D joint locations, which we refer to as delta poses, are accumulated into smooth and robust 3D trajectories and fused with direct 3D predictions. 
This principle is visualised in Fig.~\ref{fig:teaser}-(b) and can be compared with the simultaneous prediction of 3D locations and sparse 3D scene flow. 
2) Second---as events observe changes in the 2D space continuously (in the sense that their temporal resolution is substantially higher than what is required to capture the fastest human motions)---we formulate the prediction task as a continuous process, where 3D motion continuously evolves in response to asynchronous events. 
We demonstrate that state space modelling \cite{Gu2021, Zubic_2024_CVPR} is naturally suitable for continuous prediction of 3D human poses using event streams. 
3) Additional supervision with object-background segmentation masks and intermediate 2D heatmaps inherited from RGB-based vision \cite{Millerdurai2024CVPR, Millerdurai2025IJCV} can be avoided.
A neural architecture can learn to extract all intermediate features and representations necessary for 3D human pose estimation with a reduced set of pre-defined design choices. 

All these insights are reflected in our new approach, 
\textit{Event-based 3D Pose State Machine (E-3DPSM)}, 
which treats egocentric event-based 3D human pose estimation as a continuous temporal process. 
\textit{E-3DPSM} maintains a latent state that evolves with the observed motion. 
Each incoming set of events (LNES \cite{rudnev2021eventhands}) updates this internal state, which encodes the current pose, its uncertainty, and the learned spatiotemporal context.  
Predicted 3D delta changes 
correspond to event-level variations and 
(along with the evolving latent state, accumulating motion cues over time) produce temporally consistent 3D poses. 
To further mitigate drift and temporally stabilise 3D reconstruction, we introduce a learnable Kalman-filter-inspired fusion module that adaptively integrates global and delta predictions. 
In the absence of events, the latent state retains the last estimate, eliminating the need for explicit caching mechanisms used in the previous approaches \cite{Millerdurai2024CVPR, Millerdurai2025IJCV}. 
\noindent In summary, the technical contributions of this paper are as follows: 
\begin{itemize} 
\item 
\textit{E-3DPSM}, the first state machine architecture that maintains a continuous 3D human pose state aligned with the asynchronous dynamics of event streams, enabling real-time tracking at $80$Hz on our hardware (Sec.~\ref{sec:method_figure}); 
\item  
A spatiotemporal pose encoder that fuses spatial motion cues with long-range temporal dependencies, 
leveraging  
deformable attention and bidirectional state-space modelling 
to remain robust
under occlusion and rapid motion; 
\item 
A learnable 
neural module that dynamically balances direct and delta predictions, mitigating drift and ensuring smooth trajectories even under sparse and noisy events. 
\end{itemize} 

\textit{E-3DPSM} achieves state-of-the-art results on two egocentric event benchmarks, reducing MPJPE and PA-MPJPE by ${\sim}19\%$, and jitter by up to $2.7\times$ compared to the previous state of the art 
\cite{Millerdurai2024CVPR,  Millerdurai2025IJCV} (Sec.~\ref{sec:experiments}). 

\section{Related Work} 
\label{sec:related_works} 
\noindent \textbf{Egocentric 3D Human Pose Estimation.} 
Egocentric 3D human pose estimation has gained traction with the rise of VR/AR applications requiring full-body 3D motion recovery from head-mounted cameras. 
Early works such as EgoCap~\cite{10.1145/2980179.2980235} used a stereo fisheye capture setup and global pose estimation using off-the-shelf SLAM, while xR-EgoPose~\cite{9010983} and SelfPose~\cite{10.1109/TPAMI.2020.3029700} 
handled monocular settings. 
Later methods addressed calibration and improved 3D global pose estimation from HMD views~\cite{Wang_2021_ICCV,wang2022estimating} as well as whole-body pose refinement with diffusion models~\cite{wang2024egocentric}. 
To study a variety of settings with challenging human poses and self-occlusions, 
% and strong fisheye distortions, 
UnrealEgo~\cite{hakada2022unrealego} and UnrealEgo2~\cite{hakada2024unrealego2} projects introduced large-scale synthetic datasets and stereo architectures, with algorithmic improvements shown by Ego3DPose~\cite{10.1145/3610548.3618147} (a two-path network) and EgoTap~\cite{kang2024egotap} (a grid ViT). 
Further works explored rear-mounted cameras~\cite{hakada2025egorear}, fisheye-based self-supervision~\cite{liuegofish3d}, motion priors, reinforcement learning, and spatio-temporal transformers~\cite{yuan2019ego, Luo2021DynamicsRegulatedKP, 9711289, 10.1145/3580305.3599312, 9665856}. 
Despite this progress, RGB-based methods remain reliant on frame-based processing and often suffer from issues such as blur, low-light sensitivity, and high bandwidth demands.  
In contrast, event cameras offer high temporal resolution and robustness in challenging observation conditions, but require fundamentally different (and often new) neural architecture designs.   

\noindent \textbf{Event-Based 3D Human Pose Estimation.}
The first event-based methods for 3D human pose estimation used exocentric views. 
EventCap~\cite{eventCap2020CVPR} demonstrated high-speed monocular 3D motion capture by tracking a 3D human body template using a monocular event stream in a series of optimisation problems, and EventHPE~\cite{zou2021eventhpe} jointly estimated 3D poses and shapes from events using a two-stage neural network trained on the DHP19 dataset~\cite{Calabrese_2019_CVPR_Workshops}. 
EventEgo3D~\cite{Millerdurai2024CVPR} and EventEgo3D++~\cite{Millerdurai2025IJCV} introduced egocentric event-based 3D human pose estimation. 
Their architecture includes two branches, i.e., 1) an encoder-decoder to convert events into 2D heat maps and lift them to 3D poses, and 2) an event stream segmentation module to filter out background events. 
This method predominantly adopts components from RGB-based methods (i.e.,~not specifically tailored to event streams) and 
does not reach the accuracy required in many applications with HMDs (suffers from temporal jitter). 
In contrast, we rethink the problem and tailor all design choices to the egocentric event-based setting. 

The works by Gehrig et al.~\cite{Gehrig_2023_CVPR} and Zubic et al.~\cite{Zubic_2024_CVPR} proposed a recurrent ViT and a state-space model (SSM) for event-based object detection, while PRE-Mamba~\cite{Ruan_2025_ICCV} introduced an SSM for event stream deraining. 
Lang et al.~\cite{Lang_2025_WACV} used a Mamba-type SSM for context-aware fusion of RGB and event features in 3D human pose estimation. 
SSMs are also widely used in problems with RGB and point cloud inputs \cite{liu2024mamba4d}. 
Inspired by all these insights, we bring SSMs to the realm of event-based dynamic 3D vision and formulate egocentric 3D human pose estimation as a continuous dynamic process. 
By adopting a neural SSM, we explicitly model temporal evolution and inter-frame pose differences, leading to substantially improved accuracy and temporal consistency over prior event-based methods. 
This allows us to avoid event segmentation masks and intermediate 2D heatmaps \cite{Millerdurai2024CVPR, Millerdurai2025IJCV} as potential sources of inaccuracies. 

\methodfigure

\section{Preliminaries}

\textbf{Event Cameras.} 
Event cameras are bio-inspired sensors that record per-pixel brightness changes asynchronously instead of capturing full image frames at fixed rates. A pixel $(x, y)$ generates an event $e=(x, y, t, p)$ whenever the change in log intensity $I$ exceeds a threshold $C$, with polarity $p \in \{-1,1\}$ denoting the sign of the change:
\begin{equation}
\Delta I(x, y, t) = |I(x, y, t) - I(x, y, t-\Delta t)| \geq C.
\end{equation}
This results in sparse and high-temporal-resolution data, which is well-suited for fast motion and dynamic scenes. 

\noindent \textbf{LNES Representation.} Given a set of events $e_i = (x_i, y_i, t_i, p_i)$ collected within a time window of $T$ ms, we construct a Locally Normalised Event Surface (LNES) \cite{rudnev2021eventhands} $\mathbf{L} \in \mathbb{R}^{192 \times 256 \times 2}$, with two separate channels for positive and negative polarities. Each event is normalised relative to the start time $t_0$ of the temporal window, i.e.,
\begin{equation}
\mathbf{L}(x_i, y_i, p_i) = \frac{t_i - t_0}{T},
\end{equation}
so that recent events map to values close to $1.0$ and older ones decay toward $0.0$. 
This encodes both the spatial location and temporal freshness of events, yielding a dense 2D grid compatible with standard neural architectures. 

\noindent\textbf{State-Space Models (SSMs)} 
provide a structured way to model long-range temporal dependencies in sequential data. An SSM maintains a latent state $\mathbf{Z}_t \in \mathbb{R}^d$  
that evolves over time according to a linear recurrence, while mapping inputs to outputs via learnable projections: 
\begin{equation}
\mathbf{Z}_{t+1} = \mathbf{A}\mathbf{Z}_t + \mathbf{B}x_t, \quad 
\mathbf{Y}_t = \mathbf{C}\mathbf{Z}_t,
\end{equation}
where $x_t$ is the input at timestep $t$, $\mathbf{Y}_t$ is the output, and $\mathbf{A}, \mathbf{B}, \mathbf{C}$ are learned matrices. Unlike recurrent networks, SSMs use an equivalent convolutional form that enables efficient parallel training while preserving the ability to run causally at inference. Recent variants such as S4~\cite{gu2022efficiently} and S5~\cite{smith2023simplified} layers stabilise training through spectral or band-limiting constraints, making them particularly effective for long-sequence modelling tasks. 
We use event-specific S5 layers, similar to Zubic et al.~\cite{Zubic_2024_CVPR}, to capture long-term temporal context in event streams as a 3D reconstruction cue.  

\section{The E-3DPSM Approach} 
\label{sec:method_figure}

\textit{E-3DPSM} estimates temporally consistent 3D human poses from a monocular egocentric event camera equipped with a fisheye lens in three stages; see 
\cref{fig:method_figure}.
First, the incoming 
raw event stream is converted into $N$ LNES frames \(\{\mathbf{L}_t\}_{t=1}^{N}\) of length $T$ ms. 
Next, the Spatiotemporal Pose Encoder Module (SPEM, \cref{sec:spem}) extracts temporally aware, joint-specific features by combining a multi-stage convolutional pyramid, per-stage deformable attention for spatial reasoning, and event-specialised SSM (S5) layers for temporal modelling, followed by a joint-query transformer decoder that reads the deepest-stage tokens. 
Finally, the Pose Regression Module (PRM, \cref{sec:prm}) predicts a direct 3D pose \(\mathbf{P}_t^{\mathrm{D}}\in\mathbb{R}^{J\times3}\) and a delta pose \(\mathbf{P}^{\Delta}_t\in\mathbb{R}^{J\times3}\), with the number of joints $J{=}16$. 
A lightweight learned fusion module combines \(\mathbf{P}_{t-1}\) and \(\mathbf{P}^{\Delta}_t\), while using \(\mathbf{P}_t^{\mathrm{D}}\) as a global anchor, 
yielding the final temporally consistent 3D joints \(\mathbf{P}_t\). 

\subsection{Spatiotemporal Pose Encoder Module (SPEM)}
\label{sec:spem}

% Single-column figure
\encoderfigure

SPEM transforms LNES event frames into rich, temporally-aware and joint-specific representations. It is composed of a multi-stage convolutional encoder, spatially adaptive deformable attention blocks, state-space model blocks
for long-term temporal reasoning, and a joint-query transformer decoder. \cref{fig:SPEM} provides details. 

\noindent  \textbf{Convolutional Feature Encoding.} 
Each input LNES frame \(\mathbf{L}_t\in \mathbb{R}^{192 \times 256 \times 2}\) is first passed through a strided convolution, producing \(\mathbf{F}^{0}_{t} \in \mathbb{R}^{96 \times 128 \times 16}\).
The encoder then processes the features through four hierarchical stages.
At stages \(\mathbf{s}\in\{1,\dots,4\}\), we apply two residual blocks and a downsampling convolution, and obtain 
\begin{equation}
\mathbf{F}^{s}_{t} = \text{Conv}\!\Big(\text{ResBlock}^{(2)}_s\big(\text{ResBlock}^{(1)}_s(\mathbf{F}^{s-1}_{t})\big)\Big),
\end{equation}
where ``\text{Conv}'' denotes a \(3\times3\) convolution with stride 2 that reduces spatial resolution, and each \(\text{ResBlock}\) \cite{7780459} is a two-convolution residual unit with BatchNorm and SiLU \cite{ELFWING20183}.

\noindent \textbf{Deformable Attention for Spatial Reasoning.}
Inspired by recent egocentric pose estimation methods ~\cite{hakada2024unrealego2, yang2024egoposeformer} that use deformable attention \cite{zhu2021deformable} to adaptively focus on pose-critical regions under occlusions and motion, we incorporate a deformable attention block at the end of each stage to refine features at every timestep. 
We first flatten the feature map $\mathbf{F}^{s}_{t} \in \mathbb{R}^{H^s \times W^s\times C^s}$ into a sequence of feature tokens $\mathbf{T}^{s}_{t} \in\mathbb{R}^{(H^s W^s)\times C^s}$, and 
then compute
\begin{equation}
\mathbf{F}^{s}_{t}=\text{DeformAttn}\big(\mathbf{T}^{s}_{t},\ \mathbf{T}^{s}_{t},\ \mathbf{R}_s\big)\in\mathbb{R}^{(H^s W^s)\times C^s},
\end{equation}
where $H^{s}$, $W{^s}$, and $C{^s}$ are stage-dependent. The reference points \(\mathbf{R}_s\) are initialised on a normalised uniform grid and optimised end-to-end. In this setup, $\mathbf{T}^{s}_{t}$ acts as query, key, and value, so that each token attends to itself and its neighbours through learned deformable offsets. Deformable attention shifts sampling toward joint-critical neighbourhoods, allowing tokens to attend to relevant body parts under strong egocentric (fisheye lens) distortions. 

\noindent \textbf{SSM for Bidirectional Temporal Modelling.}
While spatial reasoning is handled within each frame, modelling motion over time is critical, especially under occlusions.
For this, we insert S5 layers \cite{Zubic_2024_CVPR}, a recent state-space model (SSM) variant specialised for event streams, at selected encoder stages $\hat{s} \in \{2, 4\}$ to aggregate long-range temporal context independently at every spatial location \((H^{\hat{s}}, W^{\hat{s}})\).
Let \(\mathbf{F}^{\hat{s}}_{1:N} = [\mathbf{F}^{\hat{s}}_{1},\dots,\mathbf{F}^{\hat{s}}_{N}] \in \mathbb{R}^{(H^{\hat{s}}W^{\hat{s}}) \times N \times C^{\hat{s}}}\) denote the per-location feature sequence at stage $\hat{s}$.
The SSM layer transforms this sequence and returns temporally refined features $\widetilde{\mathbf{F}}$ with an internal state $\mathbf{Z}$ as follows: 
\begin{equation}
\,\widetilde{\mathbf{F}}^{\hat{s}}_{1:N},\ \text{and}\;\,
\mathbf{Z}^{\hat{s}}_{t}
=
\mathrm{SSM}_{\hat{s}}(\mathbf{F}^{\hat{s}}_{1:N}).
\end{equation}
We use the band-limited S5 variant \cite{Zubic_2024_CVPR} (bandlimit set as $0.5$), placed after the downsampling stage.
During training, SSM layers are evaluated in parallel over full sequences using their convolutional form, and we run them bidirectionally to expose past and future context. At inference, we can switch to causal forward-only updates with recurrent state propagation by carrying internal state $\mathbf{Z}_{\hat{s}}$ across timesteps, which enables real-time deployment without future frames. 

\noindent \textbf{Joint Query Decoder}. After temporal modelling, we extract joint-specific features using a lightweight transformer decoder layer \cite{NIPS2017_3f5ee243}. We follow existing works~\cite{hakada2024unrealego2,yang2024egoposeformer} and define a set of $J$ learnable joint query embeddings $\mathbf{U} = \{\mathbf{u}_1, \dots, \mathbf{u}_{16}\} \subset \mathbb{R}^{192}$. These queries act as joint-identity tokens, allowing the decoder to consistently associate each query with a specific 3D body joint across timesteps.
 At each timestep $t$, we flatten the output from the encoder's last stage $\widetilde{\mathbf{F}}^{4}_{t} \in \mathbb{R}^{6 \times 8 \times 192}$ to memory tokens $\mathbf{M}_t \in \mathbb{R}^{48 \times 192}$, and decode joint-aware features using:
\begin{equation}
    \label{eq:TransformerDecoder}
    \mathbf{F}_{t} = \text{TransformerDecoder} (\mathbf{U}, \mathbf{M}_t) \in \mathbb{R}^{16 \times 192}. 
\end{equation} 
The decoder attends each joint query to spatial memory, enabling it to learn both joint appearance and contextual interactions between joints (e.g., elbow-wrist alignment). 
Subsequently, these final representations $\mathbf{F}_{t} \in \mathbb{R}^{16 \times 192}$ are passed to PRM (Sec.~\ref{sec:prm}) for pose prediction. 

\mpjpequalitativefigure

\subsection{Pose Regression Module (PRM)}
\label{sec:prm}
Following joint-aware spatiotemporal encoding, our PRM estimates temporally consistent 3D joint positions across all LNES frames. It consists of three components: 1) a direct pose regressor for initial predictions, 2) a delta pose regressor to track motion across time, and 3) a learnable module that adaptively fuses both estimates into a unified 3D pose. 

\noindent \textbf{Direct Pose Regression}. At each timestep $t$,  we apply a lightweight MLP head to each joint query token output by the transformer decoder $\mathbf{F}_t \in \mathbb{R}^{16 \times 192}$ to predict the 3D position of each joint directly:
\begin{equation}
    \label{eq:abs_pose}
    \mathbf{P}_t^{\text{D}} = \text{MLP}_{\text{Direct}}(\mathbf{F}_t) \in \mathbb{R}^{16 \times 3}. 
\end{equation} 
The prediction at $t=1$ initialises our fusion module, and subsequent estimates act as an anchor used to mitigate drift when necessary, as determined adaptively by the fusion module. 
This branch serves as an intermediate prediction for stabilisation rather than a standalone regression target. 

\noindent \textbf{Delta Pose Regression}.
To regress the relative offset between the current and previous frames $\mathbf{L}_t$ and $\mathbf{L}_{t-1}$, we introduce a delta pose regressor. At each subsequent timestep $t > 1$, we concatenate the current joint token $\mathbf{F}_t \in \mathbb{R}^{16 \times 192}$ with an embedding $\mathbf{E}_{t-1}$ of the previous pose, and then apply a lightweight MLP head:  
\begin{equation}
    \mathbf{E}_{t-1} = \text{MLP}_{\text{pose-emb}}(\mathbf{P}_{t-1}) \in \mathbb{R}^{16 \times 64},\,\text{and}
\end{equation}
\begin{equation}
    \mathbf{P}^{\Delta}_t = \text{MLP}_\Delta([\mathbf{F}_t ; \mathbf{E}_{t-1}]) \in \mathbb{R}^{16 \times 3}. 
\end{equation}
Since event streams encode changes rather than absolute intensities, predicting relative 3D joint displacements is often more aligned with the input modality and forms an easier regression target than absolute 3D positions. 
The deltas capture short-term motion cues that remain structured over time and are, therefore, especially helpful in situations with high-speed motion and occlusions, where absolute positions can fluctuate more significantly. 

\noindent \textbf{Learned Pose Fusion}. To obtain the final pose for $t > 1$, a naive approach would be a simple addition of the direct pose from the previous timestep with the current delta pose:
\begin{equation}
    \label{eq:naive_fusion}
    \mathbf{P}_t = \mathbf{P}_{t-1}^{\text{D}} + \mathbf{P}^{\Delta}_t. 
\end{equation}
This would lead to error accumulation over time, especially when delta estimates are noisy or when direct predictions suffer from transient uncertainty (see  App.~\ref{sec:error_accumulation}). 
To mitigate this adaptively, we introduce a learnable fusion module, implemented as a differentiable Kalman-style filter \cite{kalman1960filtering, 10.5555/3157382.3157587, kloss2021train}.
This module adaptively weighs the current delta update against the predicted direct pose to produce a more stable and accurate final pose.
At each timestep $t > 1$, the module receives: 1) $\mathbf{P}^{\Delta}_t \in \mathbb{R}^{48 \times 1}$ as a motion 
% (delta) 
update, and 2) $\mathbf{P}_t^{\text{D}} \in \mathbb{R}^{48 \times 1}$ as an observation.
We treat this process as a fusion problem, i.e.,~we fuse these two measurements to compute a corrected 3D pose estimate where we maintain a latent internal state $\mathbf{X}_t \in \mathbb{R}^{48 \times 1}$ corresponding to the current estimate of the full pose, and a covariance matrix $\mathbf{\Sigma}_t \in \mathbb{R}^{48 \times 48}$ encoding the uncertainty in the estimate. \\
\textbf{1. Motion Update (Prediction Step)}:
First, we predict the new internal state based on the previous state $\mathbf{X}_{t-1}$ and 
$\mathbf{P}^{\Delta}_t$: 
\begin{equation}
\mathbf{X}_t = \mathbf{A} \cdot \mathbf{X}_{t-1} + \mathbf{B} \cdot \mathbf{P}^{\Delta}_t. 
\end{equation}
Here, $\mathbf{A} \in \mathbb{R}^{48 \times 48}$ is a state transition matrix modelling temporal dynamics, and $\mathbf{B} \in \mathbb{R}^{48 \times 48}$ modulates the effect of 
$\mathbf{P}^{\Delta}_t$. 
We also update the prediction uncertainty:
\begin{equation}
    \mathbf{\Sigma}_{t|t-1} = \mathbf{A} \cdot \mathbf{\Sigma}_{t-1} \cdot \mathbf{A}^\top + \mathbf{Q},
\end{equation}
where $\mathbf{Q} \in \mathbb{R}^{48 \times 48}$ is the learned process noise covariance, capturing uncertainty in the motion model and delta update. 

\noindent \textbf{2. Measurement Update (Correction Step)}: Next, we incorporate the direct pose observation $\mathbf{P}_t^{\text{D}}$ to refine the predicted state.
We compute the Kalman gain $\mathbf{K}_t \in \mathbb{R}^{48 \times 48}$, which balances the confidence between the motion prediction and the observation:
\begin{equation}
    \mathbf{K}_t = \mathbf{\Sigma}_{t|t-1} \cdot \mathbf{H}^\top \cdot \left( \mathbf{H} \cdot \mathbf{\Sigma}_{t|t-1} \cdot \mathbf{H}^\top + \mathbf{R} \right)^{-1}. 
\end{equation}
Here, $\mathbf{H} \in \mathbb{R}^{48 \times 48}$ is the observation matrix (identity in our case), and $\mathbf{R} \in \mathbb{R}^{48 \times 48}$ is the learned observation noise covariance, representing uncertainty in the direct pose prediction.
We then correct the internal state using the residual, i.e.,~difference between observation and prediction: 
\begin{equation}
    \label{eq:learned_fusion}
    \mathbf{P}_t = \mathbf{X}_t + \mathbf{K}_t \cdot \left( \mathbf{P}_t^{\text{D}} - \mathbf{H} \cdot \mathbf{X}_t \right).
\end{equation}
To reflect the reduced uncertainty after fusion, we update the state covariance using the Joseph form \cite{bucy2005filtering}, which provides a numerically stable update and guarantees the positive semi-definiteness of the resulting covariance matrix:
\begin{equation}
    \mathbf{\Sigma}_t = (\mathbf{I} - \mathbf{K}_t \cdot \mathbf{H}) \cdot  \mathbf{\Sigma}_{t|t-1} \cdot (\mathbf{I} - \mathbf{K}_t \cdot  \mathbf{H})^\top + \mathbf{K}_t \cdot \mathbf{R} \cdot \mathbf{K}_t^\top.
\end{equation}
In practice, $\mathbf{A}, \mathbf{B}, \mathbf{H}$ are fixed as identities, while $\mathbf{Q}$ and $\mathbf{R}$ are learned once during training and remain constant at inference, i.e., they are not frame-dependent but shared across both the motion and measurement updates. Rather than relying on fixed noise assumptions, we learn both the process uncertainty (associated with the model update) and the observation uncertainty (associated with the direct pose prediction) in an end-to-end manner. Even though $\mathbf{Q}$ and $\mathbf{R}$ are constant at inference, learning them end-to-end provides the fusion module with calibrated priors on how much to trust delta pose changes versus direct pose estimates, thereby reducing drift and improving stability under occlusions (see Tab.~\ref{tab:avg_occlusion_only_results}). 
Rather than generic temporal smoothing, it performs reliability-aware integration of direct and delta predictions, akin to sensor fusion in classical filtering, which yields more accurate and stable 3D poses.
\mainresultstable
\subsection{Loss Functions}
To supervise our model, we employ a multi-term loss function that balances absolute accuracy, temporal coherence, anatomical plausibility, and projection consistency: 
\begin{align}
        \mathcal{L}_{\text{total}} =\;& \lambda_{\text{3D}} \mathcal{L}_{\text{3D}} + \lambda_{\Delta} \mathcal{L}_{\Delta} + \lambda_{\text{2D}} \mathcal{L}_{\text{2D}} \nonumber \\
    & + \lambda_{\text{BL}} \mathcal{L}_{\text{BL}} + \lambda_{\text{BA}} \mathcal{L}_{\text{BA}}. 
\end{align}
In our experiments, we set the loss weights as follows: 
$\lambda_{\text{3D}}{=}\lambda_{\Delta}{=}\lambda_{\text{2D}}{=}0.01$ 
and 
$\lambda_{\text{BL}}{=}\lambda_{\text{BA}}{=}10^{-3}$. 

\noindent \textbf{Delta Pose Loss ($\mathcal{L}_{\Delta}$)}. To help the model in capturing fine-grained temporal dynamics, we supervise delta pose predictions with ground-truth inter-frame joint displacements:
\begin{equation}
    \mathcal{L}_{\Delta} = \frac{1}{(N-1)J} \sum_{t=2}^N \sum_{j=1}^J \left\| \mathbf{P}^{\Delta}_{t,j} - \mathbf{P}^{\Delta^{\text{gt}}}_{t,j} \right\|^2, 
\end{equation}
where $\mathbf{P}^{\Delta}_{t} = \mathbf{P}_{t} - \mathbf{P}_{t-1}$. 
This encourages the model to learn consistent frame-to-frame motion and supports our temporal fusion strategy.

\noindent\textbf{3D ($\mathcal{L}_{\text{3D}}$) and 2D ($\mathcal{L}_{\text{2D}}$) Pose Losses.} We use mean-squared error on 3D joints and their 2D projections obtained with operator $\Pi(\cdot)$ \cite{scaramuzza2014omnidirectional} :
\begin{equation}
    \mathcal{L}_{*} = \frac{1}{NJ} \sum_{t=1}^N \sum_{j=1}^{J} 
    \left\| \mathbf{\hat{P}}_{t,j}^{\text{pred}} - \mathbf{\hat{P}}_{t,j}^{\text{gt}} \right\|^2,
\end{equation}
where $\mathbf{\hat{P}}$ is the supervision target defined as 
\[
\mathbf{\hat{P}} =
\begin{cases}
\mathbf{P}_{t,j}, & \text{for } \mathcal{L}_{\text{3D}}, \\
\Pi(\mathbf{P}_{t,j}), & \text{for } \mathcal{L}_{\text{2D}}.
\end{cases}
\]

\noindent \textbf{Bone Length Loss ($\mathcal{L}_{\text{BL}}$)}. To preserve human body proportions, we compute an L1-loss on the predicted and ground-truth bone lengths. For each bone pair $(i,j) \in \mathcal{B}$ of kinematic structure \cite{human_36M}, the bone vectors are as follows:
\begin{equation}
    \mathbf{b}_{t}^{(i, j)} = \mathbf{P}_{t,i}^{\text{pred}} - \mathbf{P}_{t,j}^{\text{pred}}, \quad \mathbf{\tilde{b}}_{t}^{(i, j)} = \mathbf{P}_{t,i}^{\text{gt}} - \mathbf{P}_{t,j}^{\text{gt}};\,\text{and}
\end{equation}
\begin{equation}
    \mathcal{L}_{\text{BL}} = \frac{1}{N|\mathcal{B}|} \sum_{t=1}^N \sum_{(i,j)\in \mathcal{B}} \left| \left\| \mathbf{b}_t^{(i,j)} \right\|_2 - \left\| \mathbf{\tilde{b}}_t^{(i,j)} \right\|_2 \right|,
\end{equation}
which stabilises joint distances, especially under occlusion.

\noindent \textbf{Bone Orientation Loss ($\mathcal{L}_{\text{BA}}$)}. To maintain anatomically plausible limb directions, we minimise the cosine distance between predicted and ground-truth bone vectors:
\begin{equation}
    \mathcal{L}_{\text{BA}} = \frac{1}{N|\mathcal{B}|} \sum_{t=1}^N \sum_{(i,j)\in \mathcal{B}} \left( 1 - \frac{\mathbf{b}_{t}^{(i,j)} \cdot \mathbf{\tilde{b}}_{t}^{(i,j)}}{\left\| \mathbf{b}_t^{(i,j)} \right\|_2 \left\| \mathbf{\tilde{b  }}_t^{(i,j)} \right\|_2}\right).
\end{equation}
This complements the bone length loss by regularising the angular configurations of limbs.

\jitterqualitativefigure

\section{Experiments}
\label{sec:experiments} 
We conduct extensive experiments on the event-based egocentric benchmarks, demonstrating the accuracy and robustness of \textit{E-3DPSM} (Sec.~\ref{ssec:main_results}). We also perform ablations to analyse the influence of each component (Sec.~\ref{ssec:ablation}). 

\subsection{Experimental Setting}
\textbf{Datasets.} 
We use two datasets: EE3D-R~\cite{Millerdurai2024CVPR}, a real-world dataset captured in a laboratory, and EE3D-W~\cite{Millerdurai2025IJCV}, a real-world in-the-wild dataset. We follow the official data splits. 
See App.~\ref{sec:dataset_preprocessing} for data pre-processing details. 

\noindent \textbf{Evaluation Metrics.} We report Mean Per Joint Position Error (MPJPE) and Procrustes-Aligned MPJPE (PA-MPJPE)~\cite{kendall1989survey} (both in mm), consistent with our baselines. To measure temporal stability, we use $e_{\text{smooth}}$
\cite{PhysCapTOG2020}  
(in mm), which compares per-frame joint displacement magnitudes: 
\begin{equation}
\begin{aligned}
e_{\text{smooth}} &= \frac{1}{(N-1)J} \sum_{t=2}^{N} \sum_{j=1}^{J} \left| \text{d}_{t,j}^{\text{gt}} - \text{d}_{t,j}^{\text{pred}} \right|,\text{ with }
\end{aligned}
\label{eq:jitter}
\end{equation}
\begin{equation}
\label{eq:jitter_eqn}
\text{d}_{t,j} = \left\| \mathbf{P}_{t,j} - \mathbf{P}_{t-1,j} \right\|_2.
\end{equation}

\noindent \textbf{Baselines and Evaluation Protocol.}
We compare against EventEgo3D \cite{Millerdurai2024CVPR}, EventEgo3D++ \cite{Millerdurai2025IJCV}, and the recent RGB-based state-of-the-art method EgoPoseFormer \cite{yang2024egoposeformer}, in which the input layer is adjusted to process LNES frames. 
To ensure a fair comparison and report $e_{\text{smooth}}$, all baselines are evaluated on continuous test sequences\footnote{originally, the baselines use a random data loading strategy}. 
We do not reset internal states in our method, consistent with its continuous
SSM formulation. 
Two inference strategies are considered: \textit{causal}, where SPEM uses only the current and past frames, and \textit{non-causal}, where it accesses all $N$ frames. 

\noindent \textbf{Implementation and Training.} We implement \hbox{\textit{E-3DPSM}} in PyTorch~\cite{paszke2019pytorch} with the event-S5 layer~\cite{Zubic_2024_CVPR}. 
Unlike previous works \cite{Millerdurai2024CVPR, Millerdurai2025IJCV},
our model requires no pre-training on EE3D-S \cite{Millerdurai2024CVPR} (synthetic data). 
The ground-truth 3D pose at the most recent event in each window of $T{=}20$ ms is used for supervision, with $N{=}40$ poses. 
Smaller $T$ improve the delta regressor’s sensitivity to fine-grained motion. 
All modules are optimised end-to-end with Adam \cite{2015-kingma}, with the batch size of $32$. 
We train for $15$ epochs on EE3D-R with a learning rate $\eta = 10^{-3}$, and then fine-tune for $10$ epochs on EE3D-W with $\eta = 10^{-4}$. 
The training is non-causal, with the SPEM having access to the full $N$-frame sequence. 
It takes 34 hours 
on four A40 GPUs, while testing is performed on one A40 in the quantitative experiments. 
3D pose update rate of $80$Hz is reached on a single A6000, and 
a more lightweight NVIDIA 3050Ti supports $52$Hz. 

\subsection{Main Results}
\label{ssec:main_results}

\avgocclusiononlytable

\cref{tab:main_results} compares our approach with prior methods~\cite{Millerdurai2024CVPR, Millerdurai2025IJCV, yang2024egoposeformer} on the EE3D-R~\cite{Millerdurai2024CVPR} and EE3D-W~\cite{Millerdurai2025IJCV} benchmarks. 
Across all three metrics, \textit{E-3DPSM} in the causal mode (supporting real-time inference) consistently outperforms existing approaches, achieving $8\%$ and $19\%$ reduction in MPJPE on EE3D-W and EE3D-R, respectively, and between $1.7\times$ and $2.7\times$ lower
$e_\text{smooth}$. 
The metrics are even slightly better for the non-causal
mode. 
While non-causal inference attains the highest overall accuracy, the causal variant performs comparably, even though the model is trained non-causally, demonstrating strong generalisation and suitability for real-time deployment. 
Qualitative results in \cref{fig:qualitative_results} further highlight these improvements: While prior methods struggle with occluded regions, particularly lower-body joints (bottom-left)---and often fail even on simple poses such as standing (bottom-right) in EE3D-W---our predictions remain consistent and anatomically plausible. 
A detailed per-joint and per-action breakdown of these results is provided in App.~\ref{sec:per_evaluation}. 

We attribute these improvements to the synergy between our delta-based motion regression and learnable fusion. PRM models 3D motion as deltas aligned with the event stream’s change-based nature, while the fusion module anchors global stability through adaptive integration with direct pose estimates. This design reduces drift, smooths trajectories, and lowers $e_{\text{smooth}}$, as visualised in the representative jitter plots for walk and kick motions (\cref{fig:qualitative_jitter_results}).  
Furthermore, the SPEM enhances robustness to self-occlusion through bidirectional SSM-based temporal modelling. The occlusion-only evaluation (\cref{tab:avg_occlusion_only_results}) shows that our method achieves the lowest average error across all occluded joints, indicating improved robustness under partial or self-occlusion. A more detailed breakdown, including end-effector analysis and evaluation protocol, is provided in App.~\ref{sec:occlusion_only}. 
Additionally,  
we compare our approach against Kalman-smoothed baselines (see App.~\ref{sec:kf_comparisons}). 
Even when the results of prior methods are additionally post-processed with Kalman filtering, our model maintains significantly lower MPJPE and $e_{\text{smooth}}$, confirming that the improvements stem from our architectural design rather than Kalman smoothing. 
Overall, 
\textit{E-3DPSM} consistently achieves state-of-the-art performance in event-based egocentric 3D pose estimation, combining high temporal consistency, robustness to occlusions, and practical real-time applicability. 

\subsection{Ablation Study}
\label{ssec:ablation}

\mainablationtable

We next perform an ablation study by disabling or modifying individual modules on EE3D-R to quantify the roles of temporal modelling, spatial selectivity, and pose fusion design, as shown in \cref{tab:main_ablations}. 

\noindent \textbf{SPEM.} 
Removing all SSM blocks results in a purely spatial baseline, which severely degrades accuracy and smoothness, while using a single SSM block (only at stage four) also degrades accuracy and smoothness, confirming the importance of early-stage temporal modelling. Disabling deformable attention further reduces performance, highlighting the need for spatial adaptivity in egocentric views. 

\noindent \textbf{PRM.}
Without the fusion module, simply adding the current delta pose to the previous pose according to \cref{eq:naive_fusion} causes severe drift and the worst overall performance, confirming that corrective fusion is essential for the accuracy of the full model. 
Using only the direct pose regressor yields poor 
$e_{\text{smooth}}$, as fine motion encoded in deltas is ignored. 
A static Kalman-style fusion improves stability, but 
underperforms our adaptive fusion, demonstrating that task-specific, learnable noise weighting is crucial for robust integration of delta and direct 3D poses 
across diverse actions. 
Our full framework achieves the best overall performance, validating the effectiveness of our design choices. We also perform additional ablations on the training strategy (App.~\ref{sec:training_ablation}), the state reset strategy (App.~\ref{sec:internal_state_ablation}), the use of different event representations (App.~\ref{sec:event_repr_ablation}), and the inference event frequency (App.~\ref{sec:inference_frequency_ablation}).

\section{Conclusion}\label{sec:discussion}

Our real-time formulation of event-based egocentric 3D human pose estimation as a continuous state evolution problem demonstrated superior results in the extensive experiments 
compared to EventEgo3D++ \cite{Millerdurai2025IJCV}. 
While MPJPE and PA-MPJPE reduced by up to $19\%$ in the causal prediction mode, 
the smoothness error 
improved by $2.7\times$ on EE3D-R and almost $1.7\times$ on EE3D-W. 
The improvements are slightly greater in the non-causal operation mode. 

The largest accuracy gains have been observed for lower-body joints in occlusion-heavy actions such as crawling and crouching, where prior methods struggled and our \hbox{\textit{E-3DPSM}} produced significantly more stable and anatomically consistent predictions. 
We attribute all these improvements 
to the avoidance of event segmentation maps and intermediate 2D heatmaps, as well as our design choices that are natural for and tailored to event streams and that were validated in the ablation study, i.e.,~bidirectional SSM (integrating long-range motion cues even when events were sparse or missing), deformable attention, learned fusion of delta and directly regressed 3D poses. 
As event-based 3D human pose estimation is a nascent research field, \textit{E-3DPSM} is not without limitations, such as sensitivity to strong occlusions and highly dynamic environments, as further discussed in App.~\ref{sec:limitations}. 
All in all, we believe the design principles and insights gained in this work can benefit many other problems in event-based 3D vision. 

\noindent \textbf{Acknowledgment.} This work was partially supported by the Nakajima Foundation scholarship. 

\vspace{20pt}
\noindent \textbf{Author Contributions.} 
MD: implementation, refinement of the concept, draft writing and editing, visualisations and the video; 
HA: supervision and draft editing; 
HR: supervision and draft editing; 
CT: lab infrastructure; 
VG: method conceptualisation, project coordination, supervision, draft writing and editing. 
{
    \small
    \bibliographystyle{ieeenat_fullname}
    \bibliography{main}
}
\clearpage
\maketitlesupplementary
\appendix

\section*{Table of Contents:}
\begin{itemize}
    \item \textbf{Appendix \ref{sec:dataset_preprocessing}}: Dataset Preprocessing
    \item \textbf{Appendix \ref{sec:error_accumulation}}: Pose Drift under Naive Fusion
    \item \textbf{Appendix \ref{sec:model_efficiency}}: Model Efficiency
    \item \textbf{Appendix \ref{sec:additional_evaluations}}: Additional Evaluations
    \item \textbf{Appendix \ref{sec:additional_ablations}}: Additional Ablations
    \item \textbf{Appendix \ref{sec:hmd_setup}}: Head-mounted Device and Real-time Demo
    \item \textbf{Appendix \ref{sec:pog_analysis}}: Past Only Gain Analysis
    \item \textbf{Appendix \ref{sec:limitations}}: Limitations
    
\end{itemize}

\section{Dataset Preprocessing}
\label{sec:dataset_preprocessing}
EE3D-R \cite{Millerdurai2024CVPR} and EE3D-W \cite{Millerdurai2025IJCV} datasets provide continuous event streams without natural frame boundaries. To make them suitable for training, we discretise the streams into fixed temporal windows of $20$\,ms. Within each window, we group events into batches of ${\approx}8 \cdot 10^3$, which provides a balanced trade-off between temporal resolution and data compactness. For each discretised segment, we generate a frame-based LNES \cite{rudnev2021eventhands} event representation that preserves polarity and temporal ordering. By creating these representations beforehand, rather than during training, we ensure consistent frame counts across the continuous streams, and this greatly improves the efficiency of data loading.

\section{Pose Drift under Naive Fusion}
\label{sec:error_accumulation}
In the naive fusion approach, the 3D pose at each timestep is obtained by adding the predicted pose from the previous timestep to the current delta pose; see \cref{eq:naive_fusion}. While simple, it leads to the accumulation of errors over time, especially when delta pose estimates are noisy or when pose predictions suffer from transient uncertainties. This error accumulation results in increasing drift, causing the predicted poses to deviate from the ground truth as the sequence progresses.
\fusionerrorfigure
As shown in \cref{fig:error_accumulation}, the MPJPE steadily increases under naive fusion, highlighting the drift. In contrast, the direct pose method fluctuates due to its reliance on independent predictions at each timestep, resulting in less consistent performance. On the other hand, our learned fusion approach remains stable and produces a consistently lower error than both the naive fusion and direct pose methods, effectively mitigating drift and preserving accuracy over time.

\section{Model Efficiency}
\label{sec:model_efficiency}
We evaluate the efficiency of our approach compared to the baselines in terms of parameter count, FLOPs, GPU memory requirement, and 3D pose update rate. As shown in \cref{tab:model_efficiency}, our 
E-3DPSM incurs moderately higher computational cost than existing (more lightweight) baselines, yet remains within the same order of magnitude and achieves real-time performance on a single NVIDIA A6000 GPU. In \cref{tab:gpumemorydetailed}, we report a detailed per-module computational requirement breakdown of our method, highlighting the primary contributors. 
Our design strikes a favourable balance that enables substantial improvements in accuracy and stability while preserving real-time responsiveness. 
It demonstrates that robustness under challenging motion and occlusion can be achieved without sacrificing deployment feasibility. 

\section{Additional Evaluations}
\label{sec:additional_evaluations}

\subsection{Comparison with Kalman-Smoothed Baselines}
\label{sec:kf_comparisons}
In our method, the Kalman filter is a learned module used for pose fusion inside the network, rather than a post-hoc smoothing step. For fairness, we also apply inference-time Kalman filtering (KF) to prior baselines, where it serves only as an external temporal smoother. This experiment reveals that our improvements cannot be attributed simply to filtering, but to the way fusion is integrated and trained within the architecture. As shown in \cref{tab:main_results_w_kf}, our approach achieves substantially lower MPJPE and smoother predictions compared to Kalman-smoothed baselines. These results confirm that the gains primarily arise from our design for learned pose fusion and temporal modelling, 
which cannot be replaced by external filtering applied after prediction. 
% rather than from 

\resultswkftable
\modelefficiencytable
\gpumemorydetailed

\subsection{Per-Joint and Per-Action Evaluation}
\label{sec:per_evaluation}
We report detailed per-joint and per-action results on EE3D-R~\cite{Millerdurai2024CVPR} and EE3D-W~\cite{Millerdurai2025IJCV} datasets.
For each dataset, we provide MPJPE and PA-MPJPE per body part together with the mean across joints, as summarised in \cref{tab:per_joint_results}. Overall, the trends mirror the aggregate findings in the main paper. Improvements are consistent across nearly all joints, with particularly large gains on distal joints such as wrists, ankles, and feet, which are challenging due to fast motion and frequent self-occlusions.

We further break down performance by action classes in \cref{tab:per_action_results} using the same metrics. The results demonstrate consistent gains across diverse activities, including dance, sports, and highly articulated motions; see Figs~\ref{fig:qualitative_wild} and \ref{fig:qualitative_real}. Notably, the improvements are most pronounced in occlusion-prone actions such as kicking, crawling, and crouching.
Additionally, the jitter plots of the end effector joints on EE3D-R (\cref{fig:qualitative_ee3dr_joints_jitter_results}) and EE3D-W (\cref{fig:qualitative_ee3dw_joints_jitter_results}) highlight the reduced jitter in these occlusion-heavy joints compared to prior methods. 
\trainingablationtable

\subsection{Occlusion-Only Evaluation}
\label{sec:occlusion_only}
To quantify robustness under occlusions, we evaluate only time steps and joints that are marked as occluded by the dataset-provided visibility masks. 
% (binary per joint: 1 visible, 0 occluded). 
We focus on end-effectors that are most susceptible to self-occlusion and fast motion: elbows, wrists, knees, ankles, and feet. For each method, we report MPJPE, MPJPE-PA, and jitter plots restricted to the occluded subset.

\cref{tab:occlusion_only_table} summarizes the per-joint results for occluded end-effectors on EE3D-R and EE3D-W. The numbers show clear and consistent gains for our approach across all end-effectors. Improvements are most pronounced at the distal joints, such as wrists, ankles, and feet, where occlusions and rapid movements typically cause large errors. This strong performance under occlusions comes from the SSM blocks used in SPEM. SSM maintains an internal latent state that evolves smoothly over time, allowing the model to integrate motion information across long temporal ranges. During occlusions---when spatial features are weak or absent---the SSM maintains and propagates a coherent motion state rather than relying solely on the current input. This temporal continuity helps preserve joint trajectories and reduce jitter for occluded joints.
Overall, the occlusion-only analysis demonstrates that our method effectively mitigates failures arising from self-occlusion, leading to more reliable pose recovery under challenging visibility conditions.

\stateresetablationtable
\section{Additional Ablations}
\label{sec:additional_ablations}

\subsection{Training Strategy}
\label{sec:training_ablation}
To assess the impact of the training strategy, we experiment with directionality and sequence length on EE3D-R, as summarised in Tab.~\ref{tab:training_ablation}. 

\noindent \textbf{Directionality.} We compare causal (forward-only) training with non-causal (bidirectional) training. In the causal training setup, SSM has access only to past observations. In non-causal training, SSM incorporates both past and future context (bidirectional) during training, and when evaluated with causal inference, achieves lower errors and smoother trajectories compared to causal-only training. This demonstrates that bidirectional context at training time helps the model learn stronger motion priors that continue to generalise even when the model is deployed in a strictly causal setting.

\noindent \textbf{Pose Sequence Length.} In this ablation, we vary the number of poses $N$ used during training to study its effect on temporal modelling. With $N = 20$, SPEM receives a limited temporal context, which reduces accuracy and increases jitter. Increasing the sequence length to $N=30$ improves accuracy and reduces jitter. Training with $N=40$ poses yields the best overall performance, indicating that longer sequences allow the model to learn richer motion dynamics and produce more stable and accurate predictions.

\subsection{Internal State Reset}
\label{sec:internal_state_ablation}

We study the effect of different inference-time state reset strategies on EE3D-R. 
Since our model maintains internal states in both the SSM blocks and the learned fusion module, one natural question is whether these states should be periodically reset to avoid drift. 
To investigate this, we evaluate three settings: 1) resetting the SSM states every $40$ frames, 2) resetting the Kalman fusion states every $40$ frames, and 3) no resets, where states evolve continuously across the entire test sequence.

As shown in \cref{tab:state_reset_ablation}, periodic resets do not yield benefits and in fact can degrade performance, either by increasing error or by reducing temporal smoothness. 
In contrast, continuous state evolution without resets achieves the best results, indicating that the model learns to regulate its internal states without the need for manual intervention. 
This analysis confirms that stability in our framework naturally arises from learned dynamics and fusion mechanisms.

\subsection{Different Event Representations}
\label{sec:event_repr_ablation}
\eventreprablationtable

We analyse the influence of different event representations on the pose estimation performance. 
Prior work by Gehrig et al.~\cite{Gehrig2019} introduced a versatile end-to-end trainable voxel-based event stream representation for learning. 
We extensively experimented with it in our framework at early and intermediate project stages and found that it leads to poor generalisation across datasets in our setting.  
We also experimented with a \textit{Learned LNES} variant of the static LNES~\cite{rudnev2021eventhands}, where each 2D entry is assigned a learnable weight based on its spatial-temporal coordinates $(x, y, t)$ and polarity $p$. 
These weights modulate local event aggregation, emphasising informative motion boundaries while suppressing noise. 
As shown in \cref{tab:event_repr_ablation}, the standard 
pre-deined 
LNES \cite{rudnev2021eventhands} yields the best accuracy and temporal smoothness. 
This suggests that structured, interpretable representations, such as LNES, provide a strong inductive bias for egocentric event-based pose estimation, achieving robustness without additional learnable overhead.  
Overall, the question of whether pre-defined or learnable event stream representations for learning are the best choices remains problem-dependent and open in the broader context of event-based vision. 

\subsection{Inference-Time Event Frequencies}
\label{sec:inference_frequency_ablation}
\eventfrequenciesablationtable

\cref{tab:event_frequency_ablation} summarises the evaluation of the robustness of our model to different event window durations ($T$) during inference, effectively varying the target 3D pose update rates. 
Although the model is trained with an event window of $T = 20$ ms (50~Hz), it maintains stable accuracy across a wide range of frequencies without significant degradation. 
This flexibility is valuable for real-world deployment, where event rates may vary due to motion dynamics or hardware constraints. 
We attribute this robustness to the event-specific S5 blocks~\cite{Gehrig_2023_CVPR}, whose learnable timescale parameters dynamically adapt to varying temporal resolutions. This capability allows the model to adjust to changes in the effective event rate, maintaining stable temporal modelling and smooth pose evolution across different inference frequencies.

\subsection{Learning Strategy for Q and R}
\label{sec:q_r_ablation}
\qrablation

We evaluate an input-dependent formulation of the process ($\mathbf{Q}$) and measurement ($\mathbf{R}$) noise covariances by predicting them with lightweight MLPs conditioned on feature embeddings $\mathbf{F}$ (see \cref{sec:spem}). Since $\mathbf{F}$ depends on the internal latent state $\mathbf{Z}$, the resulting $\mathbf{Q}$ and $\mathbf{R}$ are implicitly both input- and state-dependent. As shown in \cref{tab:q_r_ablation}, this variant performs worse than our proposed globally learned $\mathbf{Q}$ and $\mathbf{R}$. We observe that the latter act as stable, calibrated priors for temporal fusion, whereas input-dependent covariances introduce additional flexibility that leads to overfitting and less stable filtering. 

\hmdsetup
\poganalysis

\section{Head-Mounted Device and Real-Time Demo}
\label{sec:hmd_setup}

We build a head-mounted setup following the specifications of EventEgo3D++ \cite{Millerdurai2024CVPR, Millerdurai2025IJCV}, with an additional down-facing fisheye Ximea MU050CR-SY RGB camera \cite{ximea_rgb_camera} for reference views and an NVIDIA Jetson Orin Nano Super Developer Kit for portable onboard processing (see \cref{fig:hmd_setup}). 

For our real-time demo, we deploy the Jetson Orin Nano with a power bank for fully portable operation, enabling evaluation on in-the-wild sequences under low-light and fast-motion scenarios. To visualise the predicted 3D poses, we implement a lightweight client-server viewer over WebSockets, where an iPad acts as the client device streaming poses in real time (see \cref{fig:viewerdemo}). Our method operates reliably on this portable setup, achieving ${\approx}30$ Hz pose update rates on real event streams (see our video 8:10-9:36). 
When using a laptop equipped with an NVIDIA 3050 Ti carried in a backpack, our method achieves ${\approx}50$ Hz.

\limitationfigure
\section{Past-Only Gain Analysis}
\label{sec:pog_analysis}

To quantify how temporal history improves the accuracy of occluded joints in our method, we introduce and calculate the past-only gain (POG) metric, which is defined as follows. Let $t$ denote a timestep where a joint is currently occluded and has been fully visible for the previous $N=40$ frames $(t\!-\!N,\ldots,t\!-\!1)$. We define $k$ as the history length, that is, the number of most recent visible frames before $t$ that the model is allowed to use when predicting the pose at time $t$. For instance, $k=0$ means the prediction uses no temporal history, and $k=40$ means the prediction uses the last 40 visible frames preceding the occlusion. For a given joint and occluded timestep $t$, let $\text{MPJPE}_{t}^k$ denote the per-joint MPJPE when using a history of length $k$. The POG is defined as
\begin{equation}
    \text{POG}(k) = \text{MPJPE}_{t}^0 - \text{MPJPE}_{t}^k,
\end{equation}
which measures how much MPJPE is reduced at the same occluded frame when the model has access to $k$ frames of past information. A positive value indicates that temporal history improves occlusion accuracy. We compute this metric for multiple history lengths $k$. For each $k$, we evaluate predictions at the exact same occluded timesteps $t$, compute $\text{MPJPE}_t^k$, and pair it with the baseline error $\text{MPJPE}_t^0$. Averaging these paired values across all selected occlusion frames yields a POG plot for each joint.

The resulting plot \cref{fig:pog_analysis} shows positive gains for lower-body joints. The largest improvements appear for ankles and feet, which undergo frequent occlusions in egocentric settings. Increasing the history length produces progressively lower MPJPE at occluded frames, indicating that the model benefits from a richer temporal context. This confirms that our continuous state formulation effectively preserves long-range motion structure and leverages it to recover 3D human poses under severe occlusions. 

\viewerdemo
\section{Limitations}
\label{sec:limitations}
Fig.~\ref{fig:limitationfigure} shows challenging scenarios involving strong self-occlusions during crawling (\cref{fig:limitationfigure}-A), interactions with objects (\cref{fig:limitationfigure}-B), and in the presence of other humans within the field of view (\cref{fig:limitationfigure}-C). Such situations can lead to degraded 3D pose accuracy due to missing or ambiguous motion cues. In addition, abrupt illumination changes such as flickering effects (see crouching in our video 7:50-8:06) can lead to occasional temporal instability, particularly during fast and complex motions. These limitations suggest several directions for future work: Modelling occlusions explicitly and generative pose refinement could improve the plausibility of 3D poses when observations are incomplete. 

\occlusiononlyresultstable

\peractionresultstable

\perjointsresultstable

\jitterfigs

\qualitativewildfigure

\qualitativerealfigure

\end{document}